\pdfoutput=1
\documentclass[letterpaper, 10 pt, conference]{ieeeconf}
\IEEEoverridecommandlockouts
\let\labelindent\relax
\usepackage{enumitem}
\usepackage{balance}
\usepackage[font={small}]{caption}
\usepackage{subcaption}
\usepackage{array}
\usepackage{textcomp}
\usepackage{mathtools, nccmath}
\usepackage{graphicx}
\usepackage{amsfonts}
\usepackage{amsmath}
\usepackage{amssymb}
\usepackage{algorithm}
\usepackage{algorithmic}
\usepackage{hyperref}
\usepackage{tikz}
\usepackage{arydshln}
\usepackage{multirow}
\usepackage{bm}
\usepackage{epstopdf}
\usepackage{cite}
\usepackage{pifont}
\usetikzlibrary{positioning}

\newtheorem{remark}{Remark}

\begin{document}
\title{\bf Autonomous Navigation of Underactuated Bipedal Robots \\in Height-Constrained Environments
\thanks{All authors are with Hybrid Robotics Group at the Dept. of Mechanical Engineering, UC Berkeley, USA. \tt\small\{zhongyu\_li, zengjunsjtu, shuxiao.chen, koushils\}@berkeley.edu}
}
\author{Zhongyu Li, Jun Zeng, Shuxiao Chen, and Koushil Sreenath}
\maketitle
\thispagestyle{empty}
\pagestyle{empty}

\begin{abstract}
Navigating a large-scaled robot in unknown and cluttered height-constrained environments is challenging.
Not only is a fast and reliable planning algorithm required to go around obstacles, the robot should also be able to change its intrinsic dimension by crouching in order to travel underneath height-constrained regions.
There are few mobile robots that are capable of handling such a challenge, and bipedal robots provide a solution.
However, as bipedal robots have nonlinear and hybrid dynamics, trajectory planning while ensuring dynamic feasibility and safety on these robots is challenging.  
This paper presents an end-to-end autonomous navigation framework which leverages three layers of planners and a variable walking height controller to enable bipedal robots to safely explore height-constrained environments.
A vertically-actuated Spring-Loaded Inverted Pendulum (vSLIP) model is introduced to capture the robot's coupled dynamics of planar walking and vertical walking height.
This reduced-order model is utilized to optimize for long-term and short-term safe trajectory plans.
A variable walking height controller is leveraged to enable the bipedal robot to maintain stable periodic walking gaits while following the planned trajectory.
The entire framework is tested and experimentally validated using a bipedal robot Cassie. This demonstrates reliable autonomy to drive the robot to safely avoid obstacles while walking to the goal location in various kinds of height-constrained cluttered environments.
\end{abstract}
\section{Introduction} \label{sec:introduction}
Autonomous robots that can safely travel in unknown and unstructured environments with obstacles distributed in three dimensional space can be useful for various practical applications such as navigating in narrow complex spaces with varying ceiling heights in tunnels or caves or in collapsed buildings after disasters.
This motivates us to have robots that are able to crouch down and travel underneath obstacles in confined spaces.
Moreover, there are few wheeled mobile robots that are capable of changing their intrinsic dimensions while moving around obstacles due to their low Degree-of-Freedoms~(DoFs) and design limitations. This restrains their usage in complex environments. 
Bipedal robots, on the other hand, are able to dynamically change their configuration by stretching or crouching to navigate environments with height constraints.
However, bipedal robots are high-dimensional, nonlinear and underactuated systems with hybrid dynamics. 
Trajectory generation for such robots with guarantees of dynamic feasibility while safely avoiding obstacles is challenging. 
There is existing work trying to tackle these problems for bipedal robots in confined environments, such as~\cite{schulman2014motion,dai2014whole,grey2017footstep}, but these are only validated in simulated environments.
According to the best of our knowledge, navigation with height constraints for large-scaled bipedal walking robots in the real world has not been experimentally demonstrated.

\begin{figure}[t]
\centering
\includegraphics[width=\linewidth]{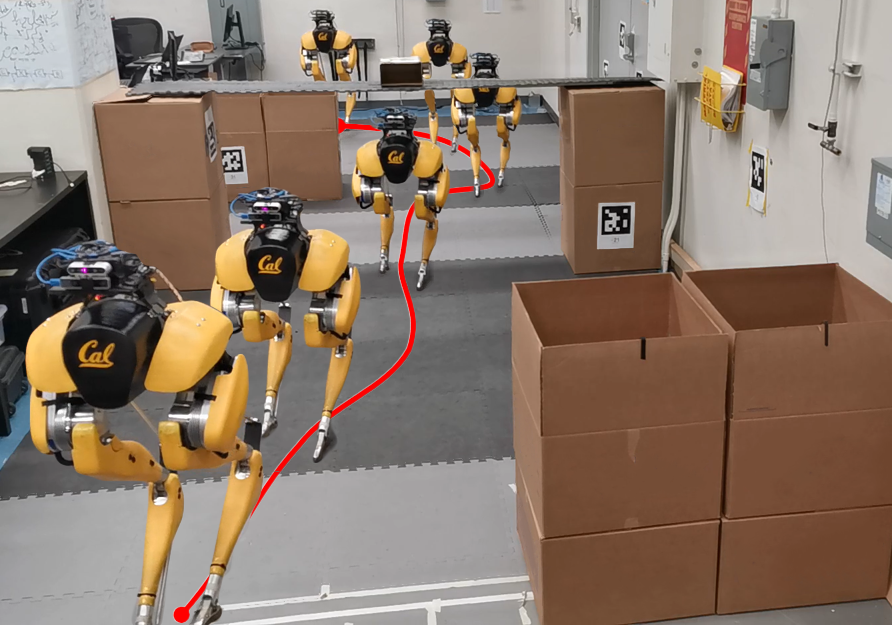}
\caption{A bipedal robot Cassie autonomously travels in congested height-constrained environments while maintaining walking gait stability and collision-free safety. The autonomous navigation framework developed in this paper enables a large-scale underactuated bipedal robot to not only navigate around obstacles but also to crouch to go under height-constrained obstacles. Experiment video can be found at \url{https://youtu.be/Da0tebC3WuE}.}
\label{fig:main-figure}
\end{figure}

In this paper, we seek to ascertain the feasibility of utilizing a dynamic bipedal walker Cassie to explore and navigate in congested and unknown environments, as shown in Fig.~\ref{fig:main-figure}.
To enable bipedal robots, like Cassie, to navigate autonomously in height-constrained spaces without prior knowledge of the environment, three major problems should be addressed.
Firstly, a collision-free path should be found and updated quickly from the current position to the goal location while exploring in the unknown space with varying admissible heights.
Next, a local trajectory for the bipedal robot is required with consideration of obstacle avoidance and robot's physical limitations. 
Moreover, the planners also need to ensure the gait stability of the underactuated bipedal robot, \textit{i.e.}, the planners should respect the underactuated dynamic coupling and prevent the robot falling and hurting nearby humans and the robot itself.
Finally, to enable the robot to reliably follow the planned motions, we need a walking controller that can change the robot's configuration, \textit{e.g.}, walking height, while maintaining stable walking gaits with different velocities. 
While these problems are challenging, a solution that addresses them will enable end-to-end navigation for bipedal robots for autonomous and safe travel in unknown height-constrained environments.

\subsection{Related Work}
\subsubsection{Motion Planning for Legged Robots:} 
Motion planning for legged robots has been an attractive topic and usually involves planning in the configuration space. 
In some prior work, paths of the root of robots are firstly found, for example, by sampling-based methods such as RRTs presented in~\cite{lavalle2001randomized, grey2017footstep, tonneau2018efficient, buchanan2021perceptive}.  
Later, robot footholds are selected over uneven terrain and the configuration of the whole body is planned by imposing foot contacts and stability constraints, such as~\cite{fankhauser2016free,hildebrandt2017real,tonneau2018efficient}. 
There are other approaches that choose to plan reachable and optimal footholds over terrains first, and then generate the whole body motion using constrained optimization, see~\cite{mastalli2017trajectory,tian2018efficient,deits2014footstep,griffin2019footstep}. 
Besides stepping over uneven terrain, some approaches also consider avoiding obstacles in confined spaces as demonstrated in~\cite{schulman2014motion,grey2017footstep,kumagai2018efficient}, but only simulated results are presented.
Recently, a hierarchical motion planning framework and experiments on quadrupedal robots to travel in height-constrained environments is presented in~\cite{buchanan2021perceptive}. 
In this work, a path of robot pose is firstly randomly sampled by RRT-Connect in~\cite{kuffner2000rrt}, followed by path smoothing while avoiding collision through CHOMP in~\cite{zucker2013chomp}. 
Footholds are selected via the method shown in~\cite{fankhauser2016free} and posture of the robot is optimized again to adapt to the planned foot placement. 
However, this last step in the pipeline ignores potential collisions, which prevents the entire framework from making collision-free guarantees.

Moreover, the above-mentioned motion planning work for legged robots only considers planning in configuration space without the dynamics of the legged robot, which results in slow and statically stable gaits. This could also lead to failures in experiments as reported in~\cite{buchanan2021perceptive}.

\subsubsection{Trajectory Optimization for Bipedal Robots:} 
More recently, a kino-dynamics model is leveraged in a Model Predictive Control~(MPC) framework to enable a quadrupedal robot to autonomously avoid obstacles in confined spaces including height constraints by~\cite{gaertner2021collision} using quasi-static gaits. 
However, collision-free navigation while considering dynamic feasibility for bipedal robots is more challenging.
This can be solved by formulating online optimization problems that enforce the bipedal robot's dynamics constraints.
Using full-order dynamic models of high dimensional bipedal robots in online optimization is computationally expensive. 
A widely-used alternative approach is through simplified models which approximate the walking dynamics, such as the Linear Inverted Pendulum (LIP) models, see~\cite{koolen2012capturability, kajita2003biped}, and Spring-Loaded Inverted Pendulum (SLIP) models, see~\cite{geyer2006compliant, poulakakis2007formal, rummel2010stable, saranli2010approximate}, based on stability criterion such as Zero Moment Point~(ZMP) in~\cite{vukobratovic2004zero} or Capturability in~\cite{pratt2012capturability}.

In some approaches, reduced-order models such as centroidal model in~\cite{dai2014whole} or Center-of-Mass~(CoM) model in~\cite{feng2015optimization} are utilized to optimize for CoM trajectories while enforcing contacts and ZMP constraints, see~\cite{herdt2010online,kuindersma2016optimization,dafarra2020non}.
Among these, \cite{dai2014whole} enables person-sized humanoid robots to travel in confined spaces, such as an obstructed door, without collision in simulation.  
However, in implementations on physical robots, these methods usually limit the robot's vertical movement, \textit{i.e.} robot's walking height is fixed to make the ZMP dynamics linear for online control, such as~\cite{kuindersma2014efficiently,kajita2003biped,feng2016robust}.
Dynamics of LIP without fixed height constraints is studied in~\cite{caron2019capturability} where a Variable-Height Inverted Pendulum (VHIP) is proposed to enable walking while changing heights for humanoids based on capturability in simulation.
However, the introduction of variable height to the LIP model is more to improve the control performance on the robot rather than to change the robot walking height significantly and enable travelling underneath obstacles. 

Cassie, which is more agile with linear feet, cannot use the above-mentioned ZMP-based approaches, see~\cite{kajita2003biped}. 
Trajectory optimization for Cassie is demonstrated in~\cite{apgar2018fast} and is formulated using MPC based on SLIP model in simulation. 
A LIP model based MPC is also developed in~\cite{teng2021toward} for discrete-time safety-critical trajectory generation proposed in~\cite{zeng2021safety} for Cassie. 
But both of them lack experiments on hardware.
MPC is also utilized in~\cite{xiong2020global} on a Hybrid LIP~(H-LIP) model of Cassie to track a simple global trajectory in experiments and to avoid nearby obstacles in simulation. 
There is a concurrent work in~\cite{huang2021efficient} on building navigation autonomy of Cassie using a RRT as a global planner and a Control-Lyapunov Function~(CLF) as a local trajectory planner.
But this work doesn't consider obstacle avoidance during the local planning.
Moreover, these works don't consider the coupled walking dynamics between planar and vertical directions and don't study autonomous navigation in confined or height-constrained spaces. 

Furthermore, almost all prior work on using reduced-order models to generate trajectories for walking robots forces the robot to mimic walking patterns computed by reduced models, like~\cite{apgar2018fast, caron2019capturability, xiong2020global}. Such an approach is effective but loses information and properties of the full-order model dynamics of the robots, thus leading to difficulties during experimental implementation on real robots, such as on Cassie as described in~\cite{apgar2018fast,xiong2020global}.
As we will see, we leverage a reduced-order model to capture the control design and its limitation on the full-order model, enabling us to achieve more agile motions during navigation on Cassie.

\subsubsection{Bipedal Locomotion Control:}
Developing a reliable locomotion controller for a bipedal robot to follow the planned footholds and configurations is critical to complete the navigation pipeline. 
There is a large body of literature to tackle this problem, see~\cite{feng20133d,kuindersma2014efficiently,caron2016zmp}. 
Hybrid Zero Dynamics~(HZD) presented in~\cite{GrChAmSi2010} offers a mathematically rigorous procedure to generate stable, fast and energy-efficient periodic gaits using the full robot model based on input-output linearization. 
Moreover, the stability of gait transition using HZD is studied in~\cite{veer2019switched} and applied to the problem of navigation by transitioning between limit-cycles and validated on bipedal walkers in simulations, such as~\cite{motahar2016composing,veer2017almost,veer2019safe}.
In experiments, HZD is validated on a range of bipedal robots with point or linear feet in~\cite{rabbit,sreenath2011compliant,da20162d}, and more recently, on Cassie in~\cite{gong2019feedback,li2020animated,reher2021control}. 
Based on~\cite{gong2019feedback} and HZD, a walking controller that is able to control a bipedal robot to crouch is firstly developed in~\cite{li2020animated}.
Variable walking height control problems are later explored by a reinforcement learning method in~\cite{li2021reinforcement} and by an H-LIP-based approach in~\cite{xiong20213d}.
In this work, we use the HZD-based controller on Cassie constructed by~\cite{li2020animated}.

\begin{figure*}[!htp]
    \centering
    \includegraphics[width=0.975\linewidth]{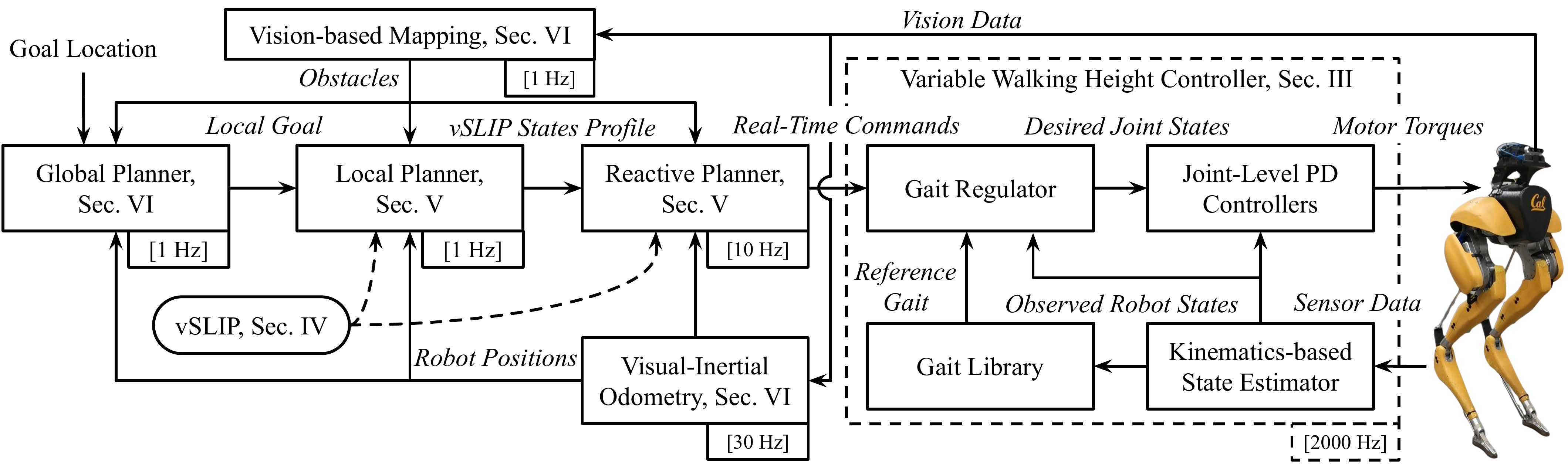}
    \caption{Our proposed autonomous navigation framework. The obstacles are perceived by a RGB-Depth camera and are registered as non-traversal areas or height-constrained areas in the map. After a goal location is set, the global planner finds a collision-free path on the map and passes a local goal to the local planner. This local planner generates a smooth trajectory to drive the robot to the local goal while avoiding obstacles. A reactive planner then tracks the planned local trajectory and gives real-time control commands to a variable walking height controller to follow the local plan. The outputs from the local planner and reactive planner also consider the dynamics of a reduced model, vSLIP, to ensure the dynamic feasibility and gait stability for the robot. Robot odometry for planners is estimated by an onboard tracking camera, and the controller is using a state estimator based on robot kinematics model, joint encoders and IMU.}
    \label{fig:framework}
\end{figure*}

\subsection{Contributions}
The primary contributions of this work are: (1) the design and development of one of the first autonomous navigation frameworks for underactuated bipedal robots to navigate unknown height-constrained environments.
In this framework, search-based A* and collocation on a reduced-order dynamical biped model work as global and local planners, respectively. 
(2) A vertically-actuated SLIP model~(vSLIP) is introduced to describe the coupled walking dynamics for variable walking heights and walking velocity and to reduce the online computation burden. 
Unlike previous attempts to force physical robots to mimic reduced-order models, we introduce a modelling method in the vSLIP to capture the properties and limitations of HZD-based locomotion control that leverages the full-order dynamics model. 
(3) Such a reduced-order model is utilized in hierarchical collocation-based trajectory planners to find trajectory profile and real-time control commands that also respect gait stability and safety constraints for bipedal robots.
(4) Using the proposed framework, we present experiments on the bipedal robot Cassie illustrating safe navigation in height-constrained environments.
This framework is one of the first to demonstrate considerable level of autonomy to enable a bipedal robot to travel without collision nor losing balance while exploring various unknown confined environments.
This proposed autonomy serves as a milestone towards real-world deployment of person-sized bipedal robots like Cassie.

\subsection{Paper Structure}
The paper is structured as follow. The overview of the proposed navigation autonomy is introduced in Sec.~\ref{sec:navigation-framework}. In Sec.~\ref{sec:cassie-control}, the Cassie experimental platform is introduced and its HZD-based variable walking height controller is briefly discussed. The gait stability constraint of this walking controller is also introduced there. Sec.~\ref{sec:vslip} develops the vSLIP model to capture variable height dynamics and limitations of the HZD-based controller, which is later utilized in optimization-based local trajectory planners developed in Sec.~\ref{sec:local-planners}. Robot localization, mapping, and global planning are presented in Sec.~\ref{sec:mapping}. Numerical validation in simulation and experimental results are presented in Sec.~\ref{sec:simulation} and Sec.~\ref{sec:experiments}, respectively. The proposed autonomy is also extended to navigating on height-constrained sloped terrain in Sec.~\ref{sec:slope}. Conclusion and future work are discussed in Sec.~\ref{sec:conclusion}.
\section{Overview of the Navigation Autonomy} \label{sec:navigation-framework}
The proposed autonomous navigation framework for the Cassie bipedal robot is illustrated in Fig.~\ref{fig:framework}. 
The core part of this autonomy is the hierarchical design of the planners.
This includes a (1) global planner to search for a collision-free path on the global map, (2) a local planner to generate dynamically-feasible trajectory to lead the robot to track a local goal location from the global planner while avoiding nearby obstacles, and (3) a reactive planner outputting real-time commands to the walking controller while providing dynamic feasibility and safety guarantees.

In this framework, the environment is perceived by a RGB-Depth camera, and the detected obstacles are classified as non-traversal areas and height-constrained areas~(Sec.~\ref{subsec:mapping-details}). The detected obstacles are registered into a 2.5D grid map as demonstrated in~\cite{kweon1989terrain} which encodes the information of admissible walking height in each grid, and the map is updated at 1~Hz. Moreover, the robot's odometry is estimated simultaneously by a tracking camera based on Visual Inertial Odometry~(VIO) as discussed in~\cite{leutenegger2015keyframe} at 30~Hz.

After being given a goal location, a global planner using A* is designed to find a path of collision-free waypoints on the map from the robot's current position to the goal~(Sec.~\ref{subsec:global_planner}).
A local goal is found on the global path within a range of the robot's current position, \textit{e.g.}, a local goal which is 1-meter ahead of the current robot position, to send to the local planner.
The global planner and the local goal are updated at 1~Hz in order to synchronize with the local planner. 

After being given the position of the local goal from the global planner, an optimization-based local planner (Sec.~\ref{subsubsec:local_planner}) uses vSLIP dynamics to compute a dynamically feasible state profile to reach the local goal.
This local planner not only respects the robot's gait stability constraints but also ensures the robot avoids obstacles. 
The resulting trajectory from this local planner lasts 3 seconds and is replanned every 1 second in this work. This local planner provides a smooth trajectory to lead the robot to reach the local goal that is 1 meter ahead. However, the 1~Hz frequency of this planner is still not fast enough for an agile dynamic walker when it is exploring an unknown and confined environment. Therefore, a reactive planner (Sec~\ref{subsubsec:reactive_planner}) that replans at higher frequency, 10~Hz, is developed.

This reactive planner uses the same optimization schematic as the local planner but has a shorter-term preview, replans at 10 Hz and outputs a short-term discrete state profile.
Its target state is chosen as 0.3~m ahead of robot's position in the planned state profile from the local planner.
It only sends the walking velocity and walking height from the closest planned state as the real-time command to the walking controller.
As we will see, this reactive planning strategy helps ensure that each control command has a guarantee of dynamics feasibility, gait stability, and collision-free safety.
Moreover, in this way, we can combine the advantages from the local planner, that is the ability to have longer preview of the robot's trajectory, and the one from a fast reactive planner that is to handle real-time disturbances and modeling error of the reduced model.

After being given the planned walking velocity and walking height, the variable walking height controller sends torque commands to the robot at 2 kHz.
Moreover, the robot states for the controller is estimated quickly based on robot kinematics model using onboard sensors such as motor encoders and IMU.
As there is no pre-built map used, this vision-to-torque framework can enable bipedal robots to safely navigate with agility in new unknown cluttered and height-constrained environments without losing balance.

\section{Cassie and Locomotion Control} \label{sec:cassie-control}

Having introduced the overall navigation framework, we next present Cassie and its variable walking height controller to better understand properties of the system to plan for. 

\subsection{Cassie Robot Model}
\begin{figure}[t]
    \centering
    \includegraphics[width=.7\linewidth]{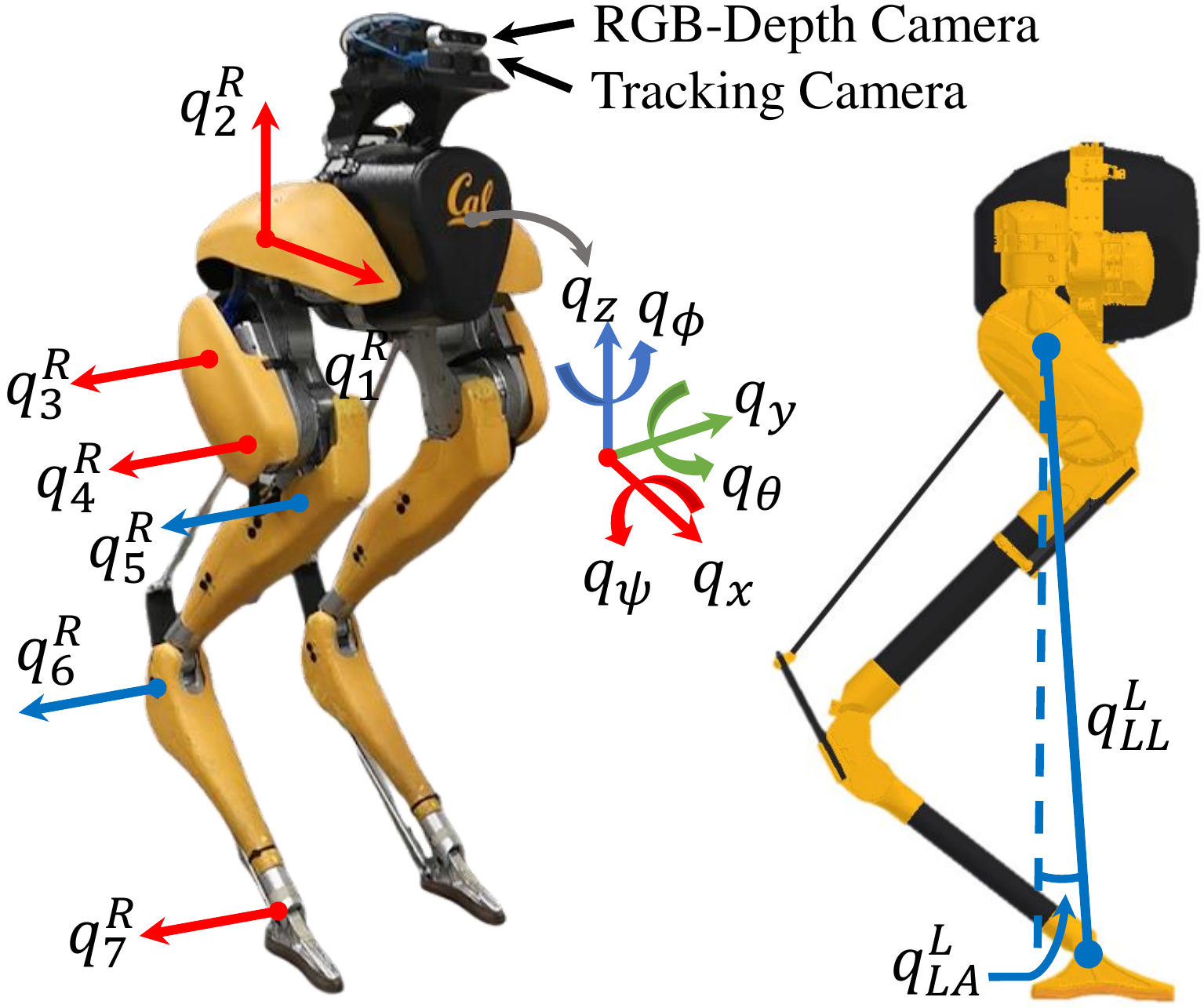}
    \caption{Cassie robot with the generalized floating-based coordinates (left) and the definition of robot virtual leg length $q^{L/R}_{LL}$ and the leg angle $q^{L/R}_{LA}$ (right). The robot is equipped with a tracking camera and a RGB-Depth camera on its top.}
    \label{fig:cassie-model}
\end{figure}
As presented in Fig.~\ref{fig:cassie-model}, Cassie is a dynamic, underactuated legged robot and has 20 DoFs. It has ten actuated motors $q_{1,2,3,4,7}^{L/R}$ and four passive joints $q_{5,6}^{L/R}$ on its Left/Right legs. Its pelvis, \textit{i.e.}, floating base, has 6 DoFs $q_{x,y,z,\psi,\theta,\phi}$ that represents sagittal, lateral, vertical transitions and roll, pitch, yaw rotations, respectively. The floating-based coordinate $q \in \mathbb{R}^{20}$ captures the joint angles and the pose of the robot. 
The underactuation in the robot is primarily due to the fact that the feet only have a pitch DOF and can't roll. Additional underactuated DOFs exist in the form of the two passive springs in each leg.
The virtual leg length and leg pitch angle are denoted by $q_{LL}(q_3,q_4)$ and $q_{LA}(q_3,q_4)$, respectively. 
Unlike its predecessor robots such as~\cite{rabbit,sreenath2011compliant}, Cassie can not only walk in the saggital direction but can also walk in the lateral direction as well as turn. 
Such an ability gives it advantages to serve as a robotic platform in navigation tasks.
Moreover, we summarize the notations and symbols we frequently use in this paper in Table~\ref{tab:notations}.

\begin{table}[]
\caption{Notations and Symbols}
\label{tab:notations}
\small
\begin{tabular}{|p{2.35cm}|p{5.3cm}|}
\hline
\multicolumn{2}{|c|}{\bf{Full-Order Model}} \\ \hline
$q \in \mathbb{R}^{20}$ & Floating based coordinates \\ \hline
$q_{x,y,z}$ & Base (pelvis) translational coordinates \\ \hline
$q_{\psi,\theta,\phi}$ & Base rotational coordinates \\ \hline
$q^{L/R}_{1,2,3,4,7}$ & Motor positions on Left/Right Leg \\ \hline
$q_{LL}, q_{LA}$ & Virtual leg length and leg angle \\ \hline
\hline
\multicolumn{2}{|c|}{\bf{Gait Library}} \\ \hline
$\mathbf{p} = [\dot{q}_x, \dot{q}_y, q_z]^T$ & Gait parameter\\ \hline
$g_{\mathbf{p}}$ & Periodic gait \\ \hline
$\mathcal{G}=\{g_{\mathbf{p}}\}$ & Gait library \\ \hline
\hline
\multicolumn{2}{|c|}{\bf{Variable Walking Height Controller}} \\ \hline
$\mathbf{c} = [\dot{q}^d_x, \dot{q}^d_y, q^d_z, \dot{q}^d_{\phi}]^T$ & Commands to the controller\\ \hline
$\mathcal{X}_{cfeas}$ & Convex feasible command set \\ \hline
\hline
\multicolumn{2}{|c|}{\bf{vSLIP Reduced-Order Model}} \\ \hline
$\mathbf{x} = [q_{x,y,z,\phi}, \dot{q}_{x,y,z,\phi}]^T$ & States \\ \hline
$\mathbf{u} = [u_{x,y,z,\phi}]^T$ & Inputs \\ \hline
$\mathbf{f}=[x_f, y_f, z_f]^T$ & Foothold position on flat ground ($z_f=0$) \\ \hline
$l(t)$, $l_0$ & Current and uncompressed leg length \\ \hline
$m$ & Total mass of the robot \\ \hline
$K(l)$ & Leg stiffness\\ \hline
\hline
\multicolumn{2}{|c|}{\bf{Trajectory Optimization}} \\ \hline
$SS$ & Safe set \\ \hline
$r_{obs}$, $r_{robot}$ & Radii of obstacles and robot footprint\\ \hline
$\underline{h}$ &  Minimum admissible height nearby\\ \hline
$s$ &  Safety margin \\ \hline
$\bm{\delta}$ & Slack variables \\ \hline
\end{tabular}

\end{table}

\subsection{Gait Library}
\label{subsec:gait-library}
To obtain dynamically-feasible walking behaviors for the bipedal robot Cassie, we generated a set of periodic walking \textit{gaits} by formulating a direct-collocation based optimization using the full dynamics model of Cassie with HZD as described in~\cite[Sec.~III]{li2020animated} based on~\cite{gong2019feedback,hereid2019rapid}.
A \textit{gait} $g$ is defined as a periodic state trajectory for all actuated joints. 
For each \textit{gait}, $\mathbf{p} = [\dot{q}_x, \dot{q}_y, q_z]^T$ represents its gait parameter.
Then, a \textit{gait library} $\mathcal{G} = \{g_{\mathbf{p}}\}$ is formulated by indexing each optimized gait with its gait parameter $\mathbf{p}$. 
We generated a gait library offline with 1331 gaits with the gait parameters as described in Table~\ref{tab:gaitlibrary}.

\subsection{Variable Walking Height Controller}\label{subsec:controller}
In the real-time controller as shown in Fig.~\ref{fig:framework}, a reference gait is obtained by linearly interpolating the gait library with respect to the observed current robot's gait parameter $\hat{\mathbf{p}}$.
A PD-based gait regulator is deployed to correct the gait in order to track the desired commands $c$. 
The control command $c$ in this work includes desired sagittal walking speed $\dot{q}^d_x$, lateral walking speed $\dot{q}^d_y$, walking height $q^d_z$, and turning yaw velocity $\dot{q}^d_{\phi}$, \textit{i.e.}, $\mathbf{c} = [\dot{q}^d_x, \dot{q}^d_y, q^d_z, \dot{q}^d_{\phi}]^T$.
This is achieved by adding regulating terms to the reference gait and the resulted regulated gait is sent to joint level PD controllers to generate motor torques to control the robot. 
More details about this controller can be found in~\cite[Sec.~IV]{li2020animated} which is extended from~\cite{gong2019feedback}.
In particular, this controller is able to regulate the robot to the command $\mathbf{c}$.

\subsection{Convex Feasible Command Set} \label{subsec:safeset}
\begin{figure}[!tp]
\centering
\includegraphics[width=1.0\linewidth]{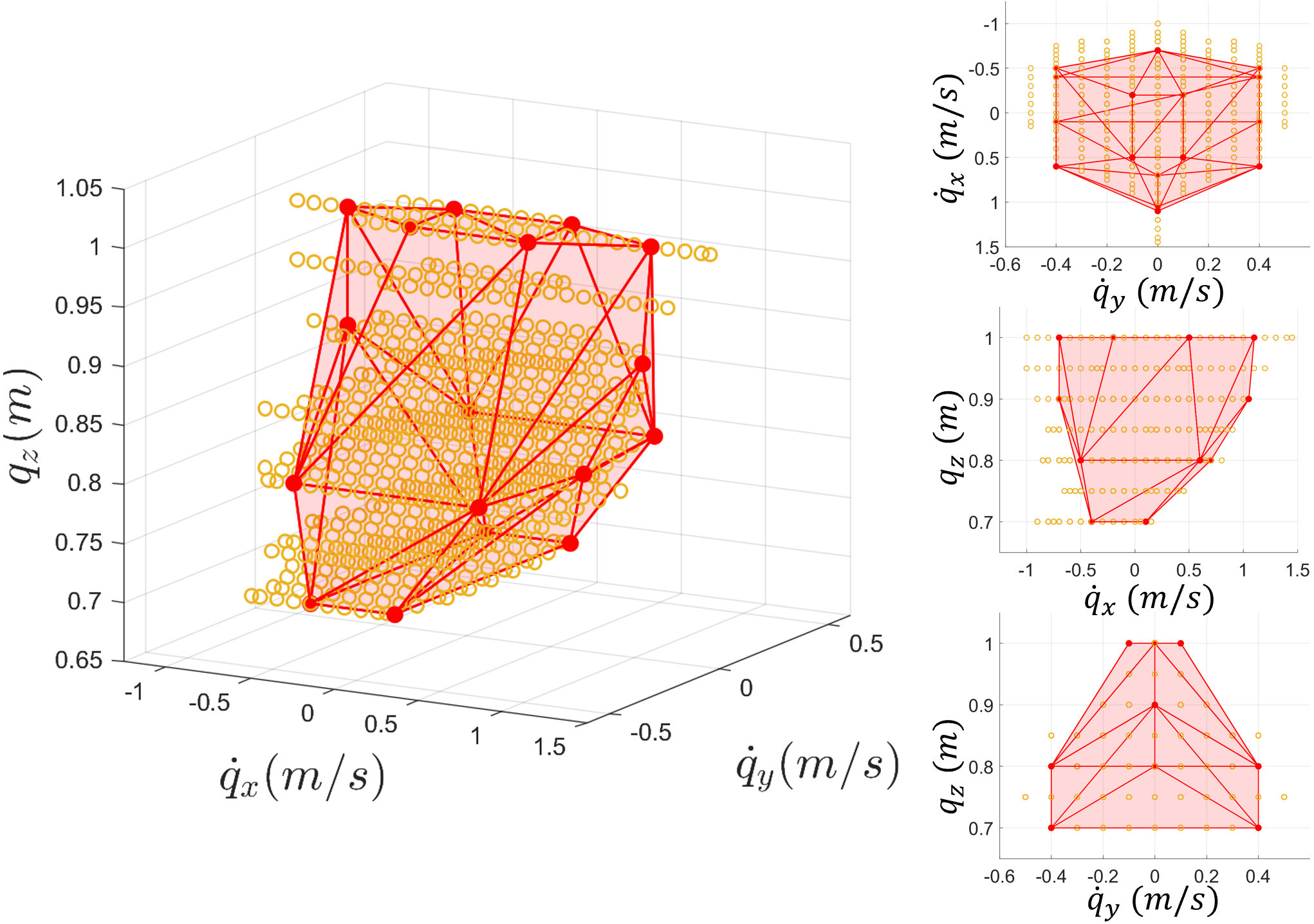}
\caption{The feasible command set (yellow) and convex feasible command subset (red). Different views of the feasible sets are shown on the right. The feasible command set consists of feasible command gait parameters and the convex feasible command set $\mathcal{X}_{cfeas}$ is its subset. Yellow circles represent different feasible combinations of walking velocity $\dot{q}_x$, $\dot{q}_y$, and walking height $q_z$. These are the commanded gait parameters that the variable walking height controller is able to stably maintain a walking gait on a flat ground in a high-fidelity simulator of Cassie. To describe those feasible commands, we obtain a convex subset denoted as $\mathcal{X}_{cfeas}$ whose vertices $\mathbf{p}_{cfeas}$ are marked as solid red points.}
\label{fig:safe_command_set}
\end{figure}

There exists varying velocity upper and lower bounds with respect to the variable walking height controller described, \textit{e.g.} when robot walks with a low walking height, it can't walk fast.
In order to quantitatively analyze such constraints, a feasible command set is constructed as shown in Fig.~\ref{fig:safe_command_set}.
This set is obtained in a high-fidelity simulation, MATLAB SimMechanics, with full-order dynamics of the Cassie walking robot.

\begin{table}[!tp]
\centering
\caption{Range of Gait Parameters used to Construct the Gait Library $\mathcal{G}$}
\label{tab:gaitlibrary}
\small
\begin{tabular}{cc}
\hline
 $\dot{q}_x$  & $\dot{q}_y$ \\ \hline
 ${[}-1,-0.8,\dots,0.8,1.0{]}$ & ${[}-0.3,-0.24,\dots,0.24,0.3{]}$ \\
\hline
 $q_z$ & number of gaits \\ \hline
 ${[}0.65,0.685,\dots,0.965,1.0{]}$ & $11\times 11\times 11=1331$  
\\
\hline
\end{tabular}
\end{table}

During each iteration in the simulation, a test command is sent to the tracking controller.
If the controller succeeds to maintain a stable walking gait on the robot for 30 seconds, \textit{i.e.} the robot doesn't fall, that input command is recorded as a feasible command.
Otherwise, that command will be considered as infeasible and not recorded.
The testing range for $[\dot{q}_x, \dot{q}_y, q_z]^T$ is between $[-1.5, -0.6, 0.7]^T$ and $[1.5, 0.6, 1.0]^T$ with a resolution of $[0.1, 0.1, 0.05]^T$ and a finer resolution at the border.
We collect all the feasible commands as the yellow circles shown in Fig.~\ref{fig:safe_command_set} and these form a Feasible Command Set. 

In this set, the maximum positive sagittal walking velocity $\dot{q}_x$ grows as the walking height $q_z$ increases, after reaching its maximum around $+1.2$~m/s, at around $0.95$~m, it drops dramatically.
The maximum negative $\dot{q}_x$ and lateral walking velocity $\dot{q}_y$ have similar trends but they reach their maximums, $-0.6$~m/s and $\pm0.5$~m/s, respectively, at $q_z=0.85$~m.
Note that the backward walking behavior in the feasible set is not symmetric to the forward walking one while side walking behavior is symmetric about the $\dot{q}_y=0$ plane. 
Moreover, the feasible walking height command range is $q_z \in [0.7, 1.0]$~m.
The range of feasible turning yaw velocity command $\dot{q}_{\phi}$ is not dependent on other dimensions, therefore, for simplicity, it is not visualized in Fig.~\ref{fig:safe_command_set}, and the range is $\dot{q}_{\phi} \in [-20, 20]$~deg/s.

Furthermore, several feasible points are selected as vertices, which are represented by solid red points in Fig.~\ref{fig:safe_command_set} to form a conservative convex subset of the original feasible command set. 
This is defined as Convex Feasible Command Set $\mathcal{X}_{cfeas}$, and its vertices are $\mathbf{p}_{cfeas}$.
As long as a given command is within this convex feasible command set, the resulting closed-loop gait obtained by the controller enforces gait stability (in the sense of not falling) on the full-order model. 
As we will see, the convex feasible command set will enable us to plan using the reduced-order model while still ensuring stability on the full-order model.

Since the HZD-based walking controller used in this work realizes periodic walking gaits, the feasible command set numerically captures the stability of the resulting periodic motions using the closed-loop controller for each command on the 3D bipedal robots. 
Since the walking period is $0.8$ second in this work and simulation time is relatively long ($30$ sec) which results in roughly $37$ full-cycle walking, the resulting motion has already become periodic if it is stable, and the robot will not fail to keep this gait during a longer test time in the absence of external perturbation.

The concept of a feasible command set was initially introduced in~\cite{li2021reinforcement} to quantitatively compare the ability of different controllers on bipedal robots. However, using the convex subset to ensure the gait stability of the command input is first introduced here.

\section{Vertically-actuated Spring-Loaded Inverted Pendulum (vSLIP)} \label{sec:vslip}
In this section, we develop a reduced-order model we call vertically-actuated Spring Loaded Inverted Pendulum~(vSLIP) to approximate the walking dynamics in planar and vertical direction for bipedal walking robots. Compared to previous reduced-order models, such as SLIP, vSLIP introduces one more actuation in the vertical direction. Moreover, the modeling of vSLIP considers the design and limitation of the robot controlled by a HZD-based controller in order to bridge the full-order model and the reduced-order model.

\subsection{Bridging the reduced-order model and full-order model}
\label{subsec:fullmodel2reduced} 
\begin{figure}[!]
    \centering
    \includegraphics[width=0.9\linewidth]{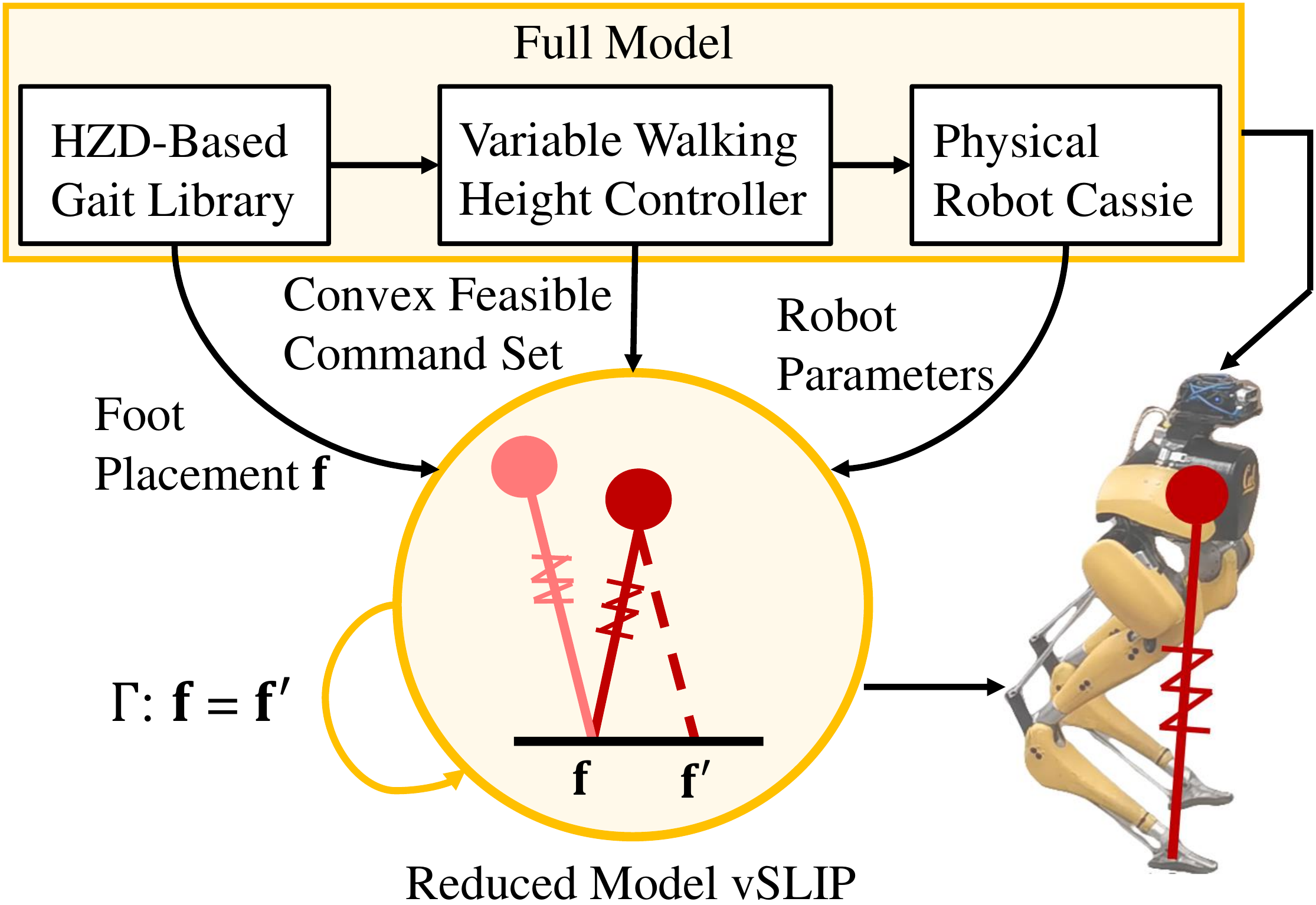}
    \caption{Proposed vertically-actuated Spring-Loaded Inverted Pendulum~(vSLIP) model. The model has one step where there is only one spring leg that supports the point mass. This model leverages the information on foot placement from the HZD-based gait library, convex feasible command set from the walking controller, and robot parameters from the physical robot.}
    \label{fig:model}
\end{figure}

Prior approaches first design reduced-order models and later develop controllers to force the robot to mimic the motions obtained from the reduced-order model such as~\cite{apgar2018fast, caron2019capturability, xiong2020global}. 
Our approach differs from this and first develops a library of dynamically feasible periodic gaits on the full-order model along with a variable walking height controller that stabilizes to a chosen periodic gait on the robot.
As illustrated in Fig.~\ref{fig:model}, we consider and utilize the restrictions coming from the full-order dynamics, the controller, and the robot in the reduced-order model. In this way, we enable the online generation of motion plans on the reduced-order model that will be feasible on the full-order model.
Specifically, this is achieved in a threefold manner: (i) We extract the foot placements from the periodic orbits in the gait library that are optimized with the full-order model;  (ii) We use the convex feasible command set describing the feasible commands to the controller, and this represents restrictions on walking velocity and walking height that result in stable gaits; (iii) We measure physical parameters of the robot. 
These are then used in the reduced-order model, vSLIP, so that this model captures the information and limitations of the full-order dynamics, the controller, and the physical robot.

\subsection{Model Representation}
\label{subsec:vSLIP}

The mathematical representation of vSLIP model is described in detail in this section. 
In vSLIP, the bipedal walking robot is simplified as a point mass supported by a massless leg with a spring and actuators during one walking step.
When one step ends and another step begins, \textit{i.e.}, transition $\Gamma$ happens, the stance leg spontaneously switches.
The state $\mathbf{x} = [q_x, q_y, q_z, q_\phi, \dot{q}_x, \dot{q}_y, \dot{q}_z, \dot{q}_\phi]^T$ represents the Cartesian 3D position and velocity of the Center-of-Mass~(CoM), and turning yaw and turning velocity of vSLIP. 
The dynamics could be defined as $\dot{\mathbf{x}}(t) = f(\mathbf{x}(t), \mathbf{u}(t))$, presented as follows:
\begin{equation}
\label{eq:dynamics}
\begin{split}
    \ddot{q}_x(t) &= \gamma(t) (q_x(t)-x_f) + u_x(t)  \\
    \ddot{q}_y(t) &= \gamma(t) (q_y(t)-y_f) + u_y(t) \\
    \ddot{q}_z(t) &= \gamma(t) q_z(t) - g + u_z(t) \\
    \ddot{q}_\phi(t) &= u_{\phi}(t)
\end{split}
\end{equation}
where $\mathbf{u} = [u_x, u_y, u_z, u_{\phi}]^T$ is the virtual input acting on the CoM and $\gamma (t) = \dfrac{K(l(t))}{m}(\dfrac{l_0}{l(t)}-1)$. Here, $m$ is the total mass, $l$ and $l_0$ are the leg length and uncompressed leg length of the reduced-order model, respectively, and $K(l)$ is the stiffness of the leg.
Moreover, the values of $m$, $l_0$ and $K(l)$ are measured from the robot.
Note that the leg stiffness parameterized as a function of the leg length is defined as below (see~\cite{xiong2018bipedal} for more details),
\begin{align}
    K(l) &= \beta_0 + \beta_1 l + \beta_2 l^2 + \beta_4 l^4 \label{eq:stiff}\\
    l &= ||[q_x, q_y, q_z]^T - [x_f, y_f, 0]^T ||_2 \label{eq:leg_length} 
\end{align}
In the vSLIP model, the walking dynamics of planar motion (xy-plane) and vertical motion are coupled by virtual leg length. 
Moreover, vSLIP has one more actuation in the vertical direction $u_z$ compared to previous SLIP models in order to change the virtual leg length significantly.
Furthermore, we decouple turning dynamics from other dimensions and simply model it as a double-integrator system. 
\begin{remark}
The way that we model the turning yaw dynamics is inspired by previous HZD-based locomotion controls in simulation and experiments where yaw doesn't affect motions in other dimensions significantly. But it does have constraint on turning velocity, \textit{i.e.}, the robot cannot change its heading orientation too fast. 
\end{remark}

In~\eqref{eq:dynamics}, the foothold position is defined $\mathbf{f}=[x_f, y_f, 0]^T$ which represents the step length $x_f$ and step width $y_f$ of vSLIP on flat ground ($z_f=0$) between two adjacent walking phases.  
The foothold for the next step $\mathbf{f}^{\prime}$ is obtained at the end of each step. 
For brevity, the superscript prime indicates the next step. 
To plan for trajectories for walking robots, the foothold position $\mathbf{f}$ is usually required to be jointly planned with model state as in prior work, see~\cite{buchanan2021perceptive,kuindersma2016optimization}.
Moreover, there exists constraints on step length and step width to make the planned foothold reachable, \textit{e.g.}, the leg cannot extend too long.
This actually brings challenges to quickly solve the planning problem. 
In the following part, we introduce the foot placement heuristics coming from the HZD-based gait library that can be utilized in vSLIP for online computation.

\subsection{Foot placement heuristics}~\label{subsec:footplacement}
In this part, we take advantage of the fact that the foothold planning and constraints are already considered during full-order model based gait optimization, and use the resulting foot placement into the reduced-order model vSLIP. In Sec.~\ref{subsec:gait-library}, the precomputed gait library optimized by the model of the robot already contains the foothold position for different gait parameters. The optimized foothold position represents the limitations on the robot's strides since the robot dynamics and physical constraints are considered during optimization. Therefore, taking advantage of the HZD-based gait library, in this work, foothold position of vSLIP doesn't need to be planned online. More specifically, the information of step length and step width is decoded from the optimized library and is utilized into the reduced-order model dynamics to fasten computational speed while ensuring the planned foothold to be reachable. Interestingly, through our study, the foot placement for next step is a function of the gait parameters $\mathbf{p}$ at the end of step, \textit{i.e.}, during transition between adjacent gaits. We denote this function as \textit{foot placement heuristics}, and it can be calculated via:
\begin{equation}
\label{eq:footplacement}
\begin{split}
    x'_{f} &= q_x^{final} + q'_{LL} \cos{q'_{LA}} \cos{q'_1} \\
    y'_{f} &= q_y^{final} + q'_{LL} \cos{q'_{LA}} \sin{q'_1}
\end{split}
\end{equation}
where $[q_x^{final}, q_y^{final}]^T$ is the state of the end of one step, and $q'_{LL}$, $q'_{LA}$, $q'_1$ are the leg length, leg pitch, and abduction, at the beginning of next gait.

\begin{figure}[!tp]
\begin{subfigure}{.32\linewidth}
  \centering
  \includegraphics[height=\linewidth]{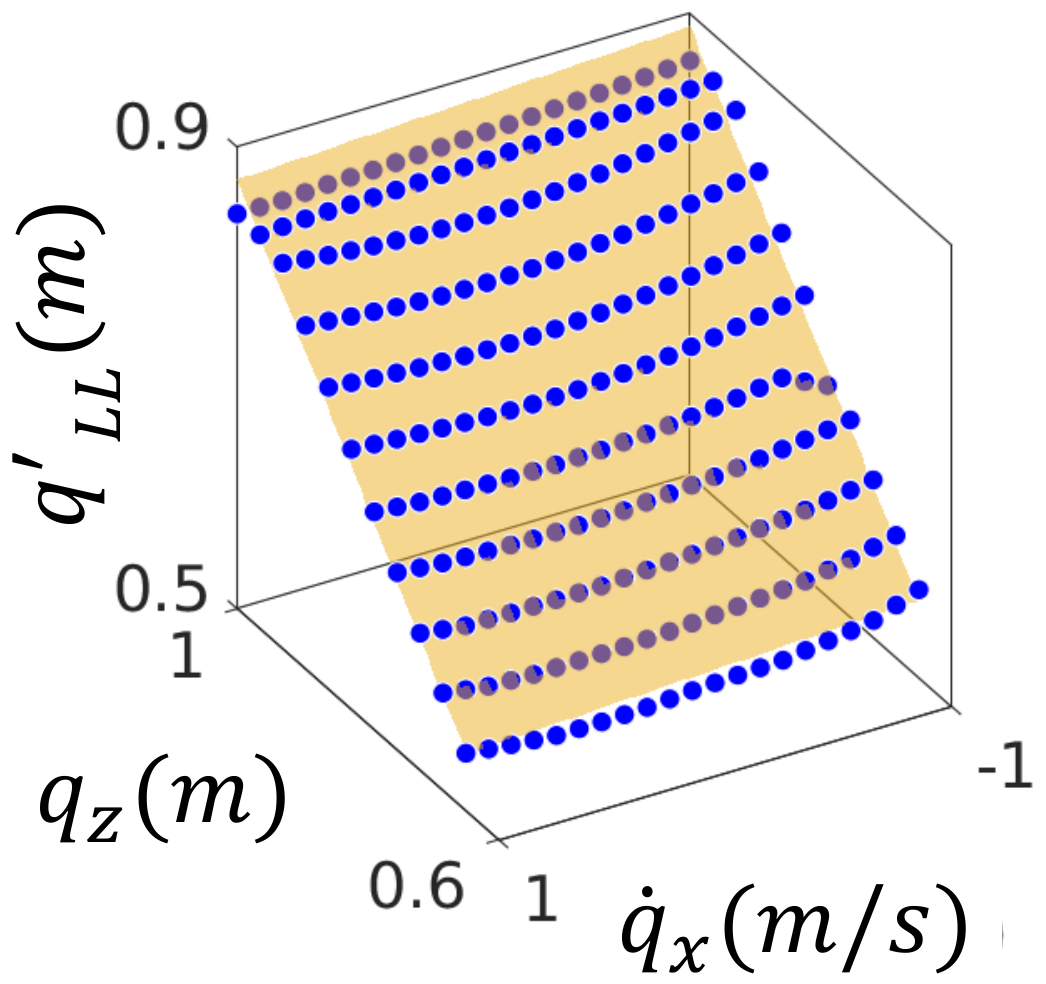}
  \caption{Leg length}
  \label{fig:touch_ground_leg_length}
\end{subfigure}
\begin{subfigure}{.32\linewidth}
  \centering
  \includegraphics[height=\linewidth]{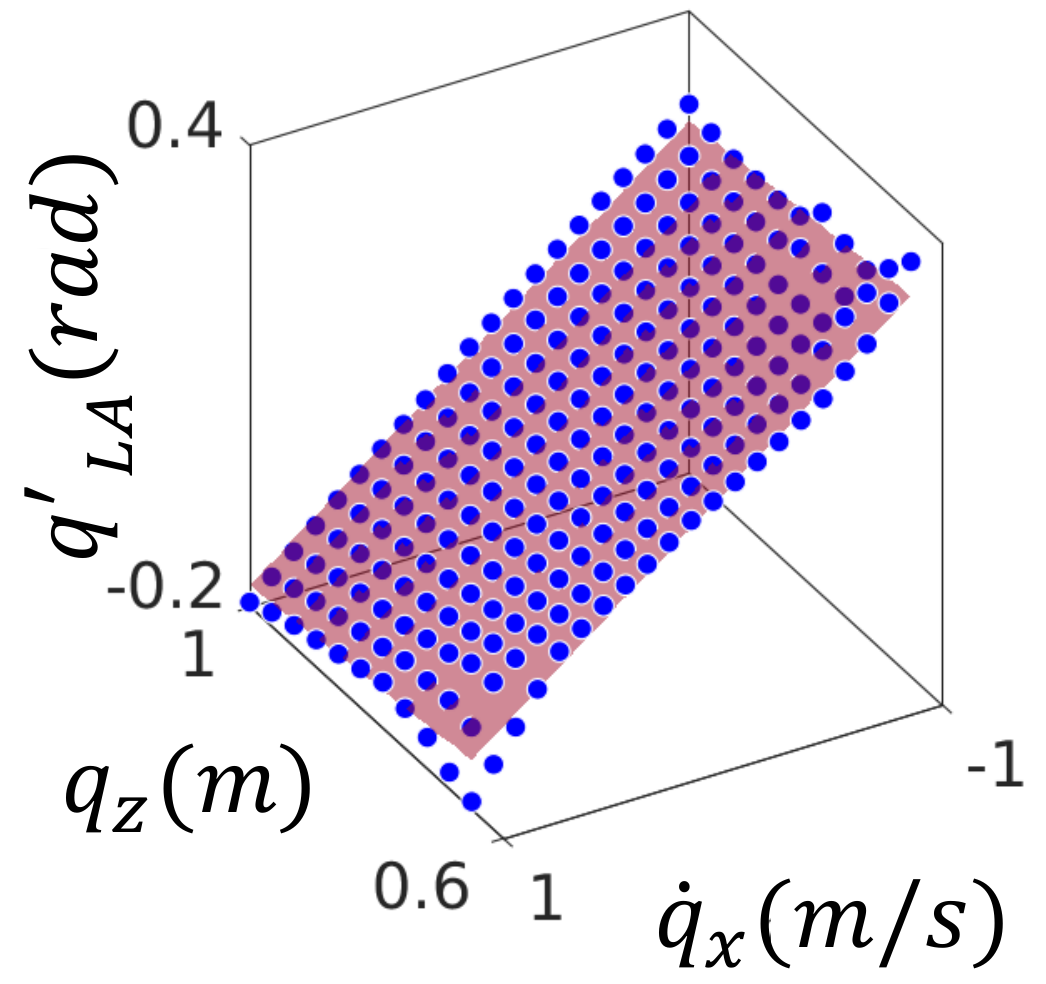}
  \caption{Leg angle}
  \label{fig:touch_ground_leg_pitch}
\end{subfigure}
\begin{subfigure}{.32\linewidth}
  \centering
  \includegraphics[height=\linewidth]{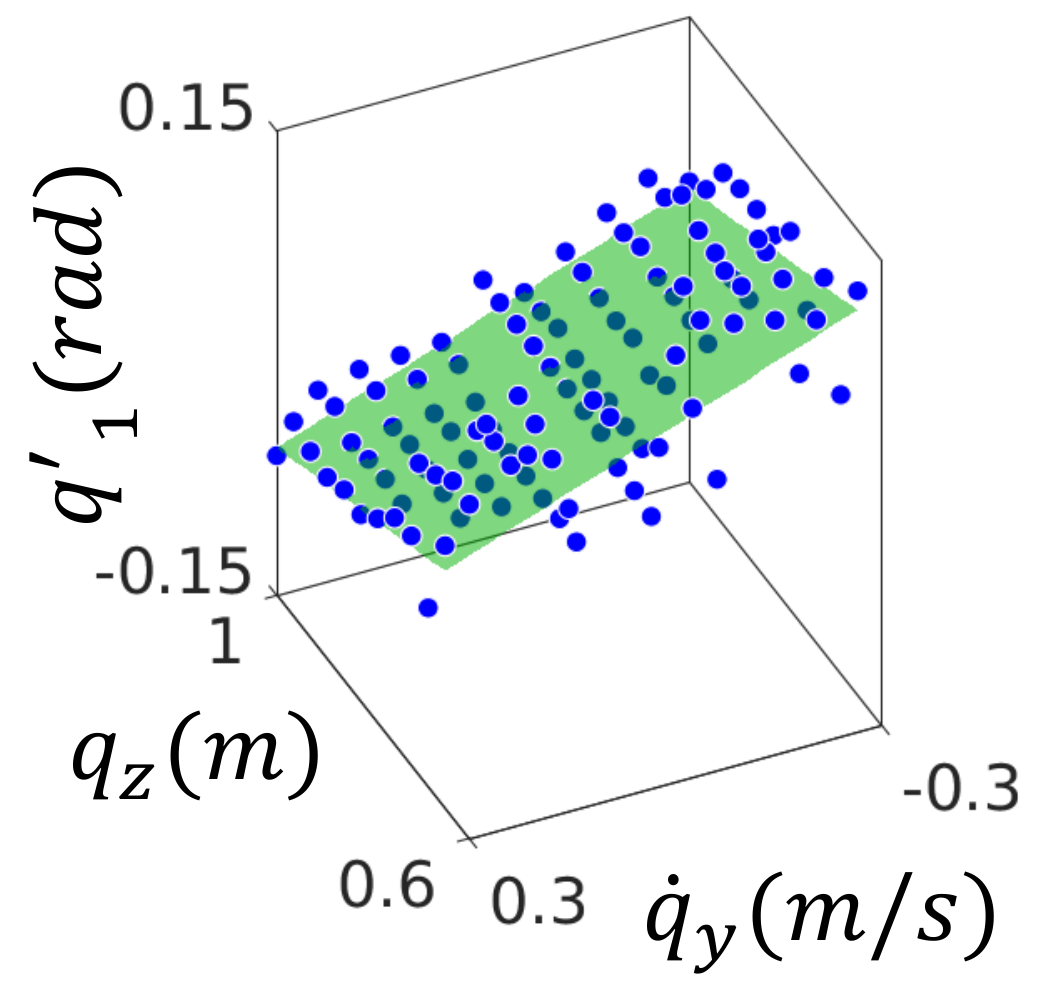}
  \caption{Leg abduction}
  \label{fig:touch_ground_abduction}
\end{subfigure}
\caption{The relationship between (a) leg length $q'_{LL}$, (b) leg angle $q'_{LA}$, and (c) leg abduction $q'_{1}$ and gait parameter $\dot{q}_x$, $\dot{q}_y$, $q_z$ at the final node. The fitting data points in this figure are from gait library and are fitted by first order polynomials, and those polynomial relations are presented by curved surfaces.}
\label{fig:footplacement}
\end{figure}

They can be formulated as explicit functions of the state of the end of one step, as shown in Fig.~\ref{fig:footplacement}, and the data points are obtained from the gait library with different gait parameters. 
To alleviate the complexity, we choose a first-order polynomial model to fit those data. 
This fitting could be described as:

\begin{equation}
\label{eq:foothold-fitting}
\begin{bmatrix}
    q'_{LL} \\ q'_{LA} \\ q'_{1}
\end{bmatrix}
= 
\begin{bmatrix}
    a_{11} & 0 & a_{13} \\
    a_{21} & 0 & a_{23} \\
    0 & a_{32} & a_{33} 
\end{bmatrix}
\begin{bmatrix}
    \dot{q}_{x} \\ \dot{q}_{y} \\ q_z
\end{bmatrix}_{final}
+ \mathbf{b}
\end{equation}
where $a_{ij}$ and $\mathbf{b}$ are the coefficients that could be solved with linear regression.
In this way, the foothold values can be updated by the gait parameters at the end of each step, presented in \eqref{eq:footplacement} and \eqref{eq:foothold-fitting}.

In conclusion, we describe the reduced vSLIP model in a manner of using the information and limitations from optimized gait library, walking controller, and physical robot based on full dynamics model. 
In return, this reduced model can be used online to generate dynamically-feasible reduced-dimensional state profile for the actual closed-loop system which is the robot controlled by a walking controller.

\section{Optimization-Based Trajectory Planners} \label{sec:local-planners}
The goal of this section is to use the vSLIP model to formulate a nonlinear program to optimize for dynamically-feasible trajectories to drive the robot to track the given local goal while respecting various constraints. 
We first introduce the formulation of collocation-based trajectory optimization on vSLIP to reach a target state. As mentioned in Sec.~\ref{sec:navigation-framework}, such a formulation is used in both a local planner and a reactive planner. After presenting the trajectory optimization problem, we also discuss and compare the local and reactive planners.

\subsection{Trajectory Optimization}~\label{subsec:traj_opti}
\begin{figure*}[!htp]
\centering
\begin{subfigure}{.4\linewidth}
  \centering
  \includegraphics[width=0.975\linewidth]{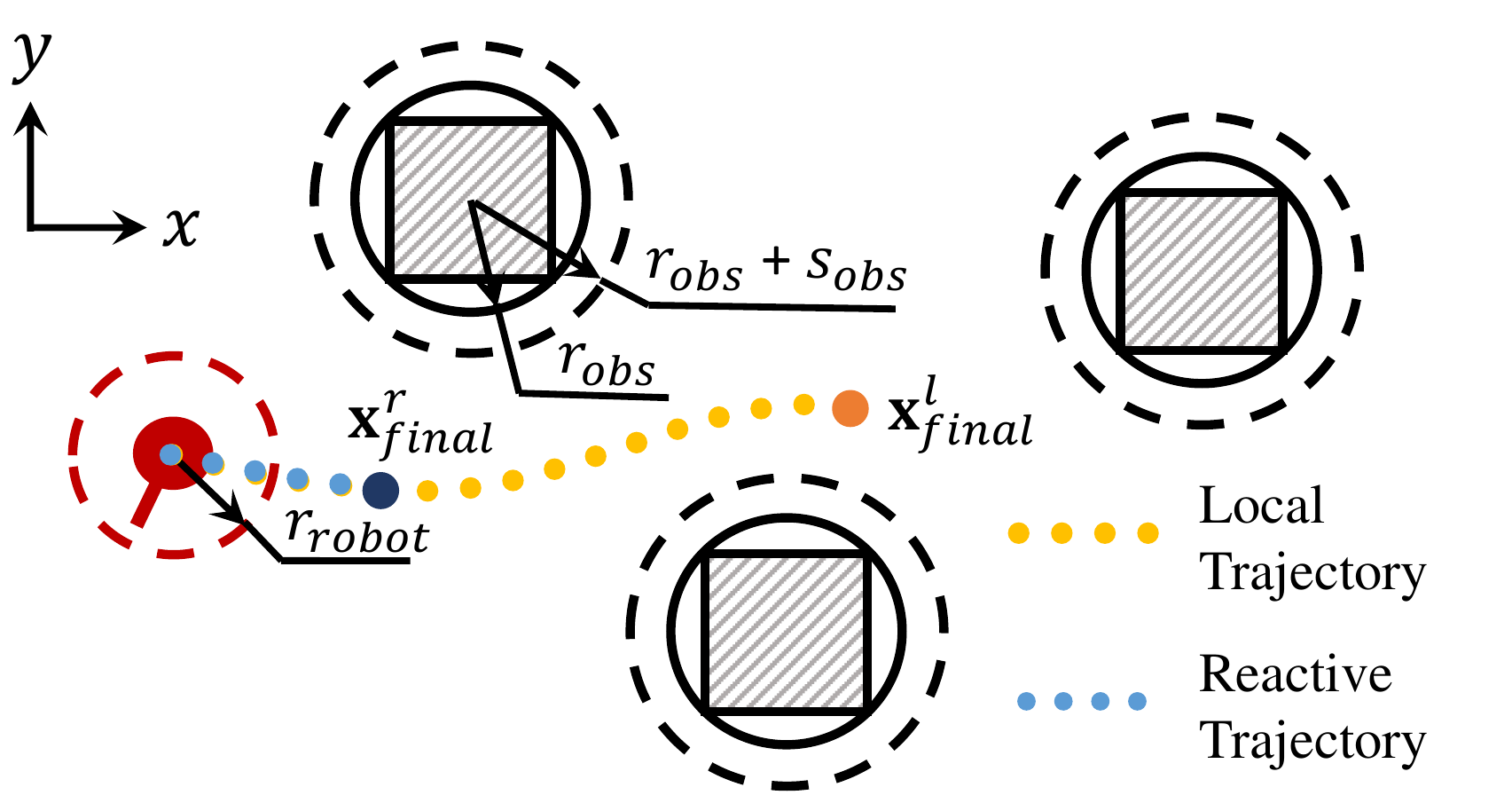}
  \caption{Obstacle avoidance constraint}
  \label{subfig:opt-obsconstraints}
\end{subfigure}~
\begin{subfigure}{.4\linewidth}
  \centering
  \includegraphics[width=.975\linewidth]{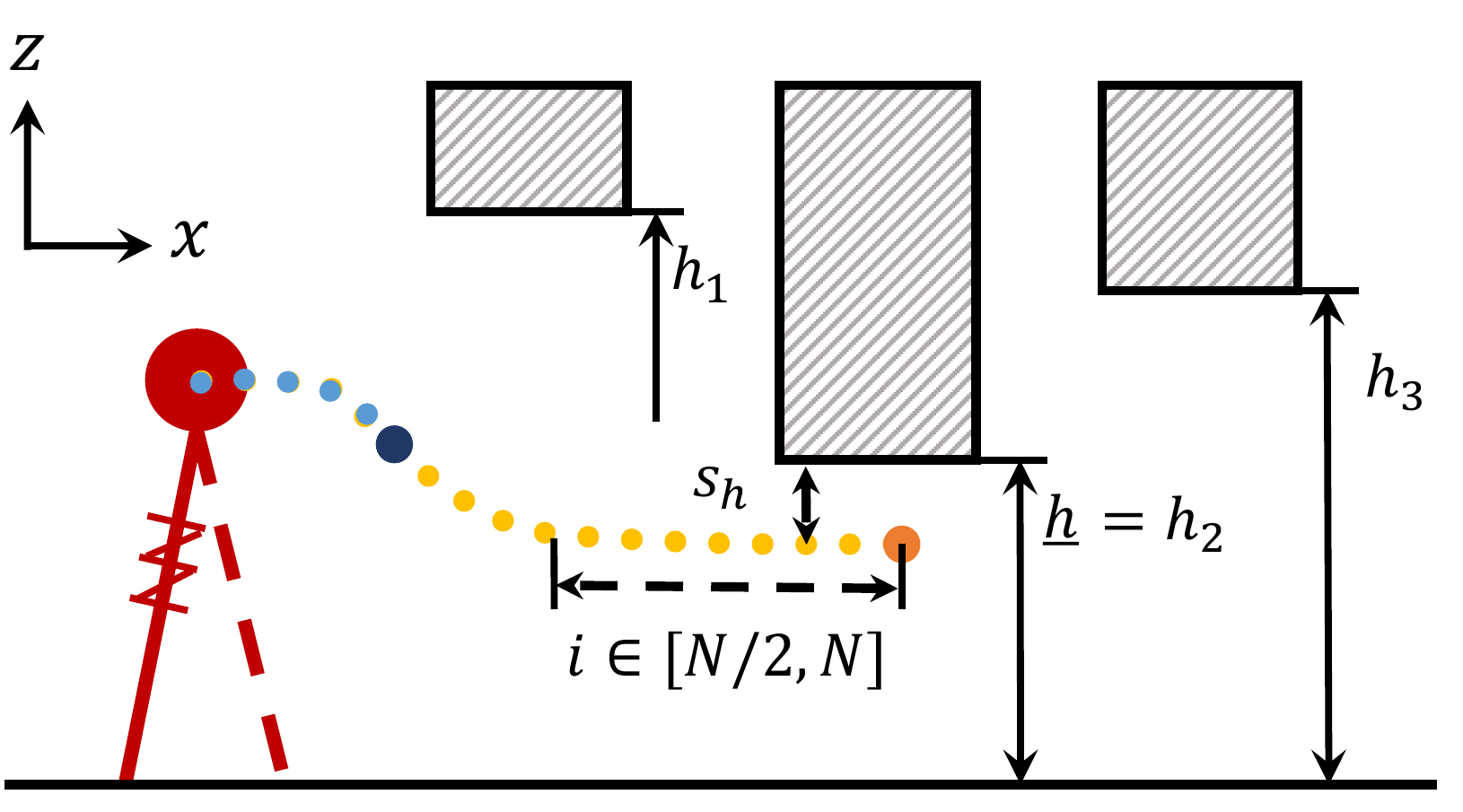}
  \caption{Admissible height constraint}
  \label{subfig:opt-heightconstraints}
\end{subfigure}
\caption{Constraints during navigation in the height-constrained environment for vSLIP. The target state for the \textit{local} planner $\mathbf{x}^l_{final}$ is set as dark yellow and target state for the \textit{reactive} planner $\mathbf{x}^r_{final}$ is marked as dark blue. Yellow dots represents the planned state profile from the local planner while the blue dots stands for the reactive planner's output. (a) The robot should avoid obstacles represented by circles with radius $r_{obs}$ with $s_{obs}$ being added to the obstacle to act as safety margin, and the robot footprint having a radius of $r_{robot}$. (b) The robot needs to crouch down and travel underneath the minimal admissible height $\underline{h}$ in the surroundings. Safety margin in height direction is included as $s_h$. Moreover, the last half of the local trajectory, \textit{i.e.}, the nodes indexing from $N/2$ to $N$, should be lower than the minimal height to lead the robot to change its configuration early.}
\label{fig:opt-constraints}
\end{figure*}

\subsubsection{Formulation}
The trajectory optimization is applied on vSLIP and is formulated using a collocation method.
States $\mathbf{x} = [q_x, q_y, q_z, q_\phi, \dot{q}_x, \dot{q}_y, \dot{q}_z, \dot{q}_\phi]^T$ and inputs $\mathbf{u} = [u_x, u_y, u_z, u_{\phi}]^T$ are defined in the vSLIP model in Sec.~\ref{sec:vslip}.
The nonlinear optimization problem is on $N+1$ discrete collocation nodes with time step $\Delta t = T/N$ and fixed time span $T$. 

Given initial and target state as $\mathbf{x}_{init}$ and $\mathbf{x}_{final}$, the collocation optimization could be formulated as follows,
\noindent
\begin{subequations}
\begin{align}
    \min_{\mathbf{x}, \mathbf{u}, \bm{\delta}} ~& J (\mathbf{x}, \mathbf{u}, \bm{\delta}),  \label{eq:problem} \\
    \mbox{s.t.}~~&\mathbf{x}_{i+1} = \mathbf{x}_i + \dfrac{1}{2} (f(\mathbf{x}_i, \mathbf{u}_i) + f(\mathbf{x}_{i+1}, \mathbf{u}_{i+1})) \Delta t, \label{eq:trapezoidal}\\
    &l(\mathbf{x}_i) \leq l_0,\label{eq:leg_length_constraint}\\ &\mathbf{x}_i \in \mathcal{X}_{adm}, \label{eq:safeset_gait_parameters}\\ 
    &\mathbf{x}_i \in \emph{SS}_i,\label{eq:safeset_collision_free} \\
    &q_{z,i} \leq \underline{h} - s_h, \forall{i\in[N/2,N]}\label{eq:height_constraints},\\
    &\mathbf{x}_{0} = \mathbf{x}_{init}\label{eq:init_constraint}, \\
    &\mathbf{x}_{N} = \mathbf{x}_{final} + \bm{\delta}_{final}\label{eq:final_constraint}, \\
    &\mathbf{u}_i \in \mathcal{U}_{adm}\label{eq:input_constraint} 
\end{align}
\label{eq:optimization_problem}
\end{subequations}
\noindent
where the cost function is designed as below,
\begin{equation}
\begin{split}
    J (\mathbf{x}, \mathbf{u}, \bm{\delta}) 
    = &\sum_{i=1}^{N-1} ( \mathbf{x}_i^T Q \mathbf{x}_i + \mathbf{u}_i^T R \mathbf{u}_i \\ 
    &+ (\mathbf{x}_{i+1} - \mathbf{x}_i)^T dQ (\mathbf{x}_{i+1} - \mathbf{x}_i)) + \bm{\delta}^T D \bm{\delta}
\end{split}
\label{eq:cost-function}
\end{equation}
where $Q \in \mathbb{R}^{8 \times 8}$, $dQ \in \mathbb{R}^{8 \times 8}$ and $R \in \mathbb{R}^{4 \times 4}$, and they are diagonal matrices with non-negative entries.
Specifically, $Q$ is the weight for state cost and has positive values on the velocity terms and zero for the position ones. 
Smoothing cost weight $dQ$ is introduced to smooth the trajectory, and input cost weight $R$ penalizes the virtual input representing the acceleration terms of the vSLIP. 
Furthermore, admissible state set $\mathcal{X}_{adm}$ and input set $\mathcal{U}_{adm}$, safe set $SS$, admissible height $\underline{h}$ and its safety margin $s_h$, initial condition $\mathbf{x}_{init}$ and target state $\mathbf{x}_{final}$ will be explained in the following section.

The time span for this optimization could cover multiple walking steps which are described in Sec.~\ref{subsec:vSLIP}. At the end of each step, the foothold $\mathbf{f}$ in the next step will be updated by the foot placement heuristics~\eqref{eq:footplacement}.

Moreover, $\bm{\delta}$ is defined as $\bm{\delta}=[\bm{\delta}^T_{cfeas}, \bm{\delta}^T_{obs}, \bm{\delta}^T_{final}]^T$. 
This represent a vector of slack variables used in the constraints, and diagonal matrix $D$ has all positive entries and is sufficiently large to minimize the slack.

\subsubsection{Constraints}\label{subsec:constraints}
The constraints that the optimization problem~\eqref{eq:optimization_problem} enforces during the navigation in height-constrained environments mainly contains the state and input bounds, initial and final state conditions, dynamic feasibility, avoidance for both ground obstacles and overhanging obstacles, and gait stability of the robot.
These can be summarized as follows.

\paragraph{Dynamic feasibility} 
The vSLIP dynamics in \eqref{eq:dynamics} are imposed in the planned trajectory by trapezoidal collocation constraints between adjacent nodes via~\eqref{eq:trapezoidal}.

\paragraph{Leg length} 
The leg length of vSLIP defined by \eqref{eq:leg_length} should be always less than or equal to the uncompressed leg length $l_0$. This is realized by \eqref{eq:leg_length_constraint}.

\paragraph{States bounds and gait stability}
The admissible state set $\mathcal{X}_{adm}$ in \eqref{eq:safeset_gait_parameters} is the intersection of two sets: states within upper and lower bounds and states within the convex feasible command set $\mathcal{X}_{cfeas}$ introduced in Sec.~\ref{subsec:safeset}. The admissible state set is
\begin{equation}
    \mathcal{X}_{adm} = \{\mathbf{x}~|~\mathbf{x}_{l} \leq\mathbf{x} \leq \mathbf{x}_{u}\} \cap \mathcal{X}_{cfeas}. 
\end{equation}
Since $\mathcal{X}_{cfeas}$ is a convex set, it can be described by its $S$ vertices $\mathbf{p}_{cfeas}$ (Fig.~\ref{fig:safe_command_set}) such that
\begin{equation}~\label{eq:feasible_set_constraint}
    \mathcal{X}_{cfeas} = \{\mathbf{x}~|~\mathbf{p} = \sum_{j=1}^{S} \lambda_{j} \mathbf{p}_{cfeas,j}, \sum_{j=1}^{S} \lambda_{j} = 1\}. 
\end{equation}
The above constraint at the $i^{\text{th}}$ node can then be rewritten as follows,
\begin{subequations}
\begin{align}
    \mathbf{p}_i &= \sum_{j=1}^{S} \lambda_{ij} \mathbf{p}_{cfeas,ij}, \\
    \sum_{j=1}^{S} \lambda_{ij} &= 1 + \delta_{cfeas,i}, \\
    \lambda_{ij},~\delta_{cfeas,i} &\geq 0
\end{align}\label{eq:safeset}
\end{subequations} 
where $\lambda_{ij}$ are Lagrangian multipliers. 
If the optimizing states meet~\eqref{eq:safeset}, the walking controller will be able to maintain a stable walking gait on the robot since the command is feasible for the controller as illustrated in Sec.~\ref{subsec:safeset}.
A slack variable vector $\bm{\delta}_{cfeas}$ is added to allow small violations to ensure the feasibility of the optimization problem.

\paragraph{Collision-free safety}
The robot should be able to avoid obstacles along the trajectory and this is achieved by limiting all states in the safe set $\emph{SS}_i$ of $i^{\text{th}}$ node in \eqref{eq:safeset_collision_free}, as illustrated in Fig.~\ref{subfig:opt-obsconstraints}. 
This safe set is formulated as follows, 
\begin{align}
\begin{split}
    \emph{SS}_i = \Big\{&(q_{x,i},q_{y,i}): \forall{j}, r = r_{obs,j}+r_{robot}+s_{obs,j}, \\
    & \|\frac{q_{x,i} - x_{obs,j}}{r}, \frac{q_{y,i} - y_{obs,j}}{r}\|^4_4 \geq 1 -\delta_{obs,j}, \\
    & \delta_{obs,j} \in \mathbb{R}^{+}\} \\
\end{split}
\end{align}
where $(x_{obs,j},y_{obs,j})$ represents planar position of the detected $j^{\text{th}}$ obstacle and $r_{obs,j}$ encodes that obstacle's size.
$r_{robot}$ stands for the robot shape. 
In this case, it is set to $0.5$~m which is Cassie's footprint size. 
$l_4$ norm is used to better describe the shape of rectangle obstacles.
The constant $s_{obs}$ is set to be a safety margin for obstacle avoidance.
Another vector of slack variable $\bm{\delta}_{obs}$ is included to attain the feasibility of the optimization problem. 
However, this slack variable should be always larger or equal to zero to avoid collision. 

\paragraph{Admissible height}
In the environment with varying height constraints, there may exist different admissible heights in the robot surrounding, as shown in Fig.~\ref{subfig:opt-heightconstraints}.
Therefore, a conservative way to apply the height constraint is to set the height of the robot in the last half of the trajectory to be less than the minimal admissible height $\underline{h}$ in the local region as given by~\eqref{eq:height_constraints}.
Moreover, a safety margin of the robot height $s_{h}$ is included. 
Enforcing this constraint in the last half of the trajectory could guide the robot to lower its height early to have a smooth height transition and to prevent the robot crouching down too slow to avoid colliding with the ceiling obstacles. 

\paragraph{Initial and final states}
The starting node $\mathbf{x}_0$ should be equal to the current observed robot state $\mathbf{x}_{init}$, enforced by~\eqref{eq:init_constraint}.
Similarly, the terminal node $\mathbf{x}_N$ should also stay close to the given target state $\mathbf{x}_{final}$, enforced by~\eqref{eq:final_constraint}. 
But the target state may not satisfy other constraints, \textit{e.g.}, the target state is very close to the obstacles or is unreachable for the robot dynamics in the given time span. 
Therefore, a vector of slack variable $\bm{\delta}_{final}$ is added to adjust the distance between $\mathbf{x}_N$ and $\mathbf{{x}}_{final}$ to ensure the feasibility of the optimization.

\paragraph{Input bounds}
Equation~\eqref{eq:init_constraint} restricts the virtual input to stay in the admissible input set $\mathcal{U}_{adm} = \{\mathbf{u}~|~\mathbf{u}_l \leq \mathbf{u} \leq \mathbf{u}_u\}$. 
This could help to bound the acceleration along the trajectory, which could also smooth the optimized trajectory as $\mathbf{u}$ contains acceleration of the vSLIP. 
Please note that the inputs $\mathbf{u}$ are virtual and represent acceleration terms of the vSLIP model in~\eqref{eq:dynamics}. The actual commands given to the locomotion controller are functions of the states $\mathbf{x}$. The input bounds here are redundant terms and can be a very large range. These are to prevent unbounded acceleration in the optimized state trajectory if the stage costs are all zeros ($Q, dQ, R=0$).

\subsection{Local planner and reactive planner}\label{subsec:local_reactive_planners}
As explained in Sec.~\ref{sec:navigation-framework}, the local planner and reactive planner share the same optimization schematic explained above but serve different proposes: the local planner has longer preview to generate a smooth trajectory to lead the robot to reach the local goal location, while the reactive planner has a shorter preview but replans in a high frequency to obtain real-time commands for the controller.  
The output from the local planner is marked as the yellow trajectory in Fig.~\ref{fig:opt-constraints} and the plan from the reactive planner is shown as the blue trajectory in Fig.~\ref{fig:opt-constraints}. 
The comparison between these two planners is illustrated in Table~\ref{tab:planners_comparison} and is explained in detail as follows.

\begin{table}[!tp]
\centering
\caption{Comparison between Trajectory Planners}
\label{tab:planners_comparison}
\small
\begin{tabular}{ccc}
\hline
               & Local Planner      & Reactive Planner        \\ \hline
Num. of steps & $6$             & $1$               \\
Time span $T$  & $3$~s           & $0.5$~s           \\
Horizon $N$    & $6\times6=36$   & $6\times1=6$        \\
Apply~\eqref{eq:height_constraints}   & Yes             & No  \\
Target State    & $\mathbf{x}^{l}_{final} =[q_{x,y,z,\phi}^{l},\mathbf{0}]^T$  & $\mathbf{x}^{r}_{final}$\\
\hline
\end{tabular}
\end{table}

\subsubsection{Local Planner}\label{subsubsec:local_planner}
The role of the local planner is to generate a smooth local trajectory which complies to constraints to lead the robot to the given local goal that is 1 meter ahead.
As shown in Table~\ref{tab:planners_comparison}, this planner uses a preview with $6$ walking steps elapsing 3 seconds, and replans around every 1 second.
Such long-term preview brings benefits to the robot with longer prediction and therefore enables the robot to proceed to avoid obstacles and change walking height early with a larger safety margin.
For example, as illustrated in Fig.~\ref{fig:opt-constraints}, the height constraint~\eqref{eq:height_constraints} is enforced on the last half of the planned local trajectory.
This prevents the robot from being too late to crouch down and to avoid the height constraint. 
The target state $\mathbf{x}^{l}_{final}$ is set to $[q_x^{l},q_y^{l},q_z^{l},q_\phi^{l},\mathbf{0}]^T$, as marked as dark yellow in Fig.~\ref{subfig:opt-obsconstraints}, where the desired position terms is the given local goal location, and the velocity term is zero. 
The given local goal is selected as one-meter ahead of the robot's current position, and is found on the global path.
Moreover $q_z^{l}$ is set to 1~m to encourage the robot to recover to normal walking height (1~m) if possible. 
As this 3-second local trajectory will be updated every 1 second, robot will only follow less than half of the trajectory. 
The second half of the local trajectory with zero terminal velocities serves as a backup plan to decelerate the robot and step in place in case of a planning failure.

However, this planner suffers from a drawback. 
The robot may deviate from the planned trajectory during the 1-second update due to tracking errors of the controller or due to environmental perturbations.  
Especially, it is risky to allow person-sized bipedal robots to follow a path without any feedback and correction for 1 second.
Therefore, a reactive planner running at a higher frequency is needed.

\subsubsection{Reactive Planner}\label{subsubsec:reactive_planner}
The reactive planner is designed to run at 10 Hz in real time to enable the robot to track the planned local trajectory. 
The fast computational speed is achieved by only planning for a single step with a horizon length of 6.
Besides the horizon difference, the reactive planner has several variances from the local planner. 
Firstly, as the planned reactive trajectory is short, satisfying the height constraints for half of the horizon may not be feasible. 
Therefore, as mentioned in Table~\ref{tab:planners_comparison}, \eqref{eq:height_constraints} is not enforced in this planner.
Secondly, the target state for the reactive planner $\mathbf{x}^{r}_{final}$ is chosen from the local trajectory with a range of $0.3$~m, and not only includes target position but planned velocity, as shown in Fig.~\ref{subfig:opt-obsconstraints}.
Apart from these two changes, the reactive planner shares the same constraints and costs as the local planner, ensuring the reactive plan to meet all the constraints.
The output from the local planner is a trajectory with 6 nodes, but it only passes the next state to the controller as command $\mathbf{c}=[\dot{q}^d_x, \dot{q}^d_y, q^d_z, \dot{q}^d_\phi]^T=[\mathbf{x}_1(5),\mathbf{x}_1(6),\mathbf{x}_1(3),\mathbf{x}_1(8)]^T$ where the subscript 1 represents the first node after the initial node 0.

\subsubsection{Initial Condition}
If the initial condition $\mathbf{x}_{init}$ is set to the observed current robot state $[\hat{q}_{x,y,z,\phi}, \hat{\dot{q}}_{x,y,z,\phi}]^T$, the reactive planner is a nonlinear model predictive controller~(NMPC). 
However, due to several factors such as the vSLIP only approximates the robot dynamics and the measured velocities are very noisy on the walking robot, NMPC doesn't work well in the experiments on Cassie. 
Therefore, the initial condition is set to a combination of feedback and feedforward terms: the position part except $q_z$ in the $\mathbf{x}_{init}$ is measured while the rest is the last timestep output from the reactive planner, and the local and reactive planners share a same initial condition. 

\subsubsection{Solving the Optimization}
The optimization problems for the long-term planner and reactive planner are formulated using CasADi in~\cite{Andersson2018} and solved by IPOPT in~\cite{biegler2009large} with the initial guess linearly interpolated between the given initial condition $\mathbf{x}_{init}$ and target state $\mathbf{x}_{final}$.

In this manner, we obtain the major part of the proposed autonomy which is the pipeline to go from a given local goal to real-time control commands for walking robots. 
This is realized by using the cascading trajectory planners to optimize for a local plan with long preview and a reactive plan with fast replanning. 
The trajectory optimization schematic used in these two planners leverages vSLIP to ensure dynamic feasibility, collision-free safety, and gait stability for bipedal robots, and can be utilized online.
\section{Localization, Mapping \& Global Planner} \label{sec:mapping}
After the development of trajectory planners for Cassie, the infrastructure of the navigation autonomy, such as perception and localization of the robot, world representation via a 2.5D map, and the global planner to provide a path to reach the given goal, is introduced in this section.

\subsection{Perception and Localization}
Cassie's perception is based on robot vision, and it is equipped with one tracking camera~(Intel RealSense T265) and one RGB-Depth camera~(Intel RealSense D435i). 
Both of them are fixed on the top of Cassie's pelvis, as shown in Fig.~\ref{fig:cassie-model}. 
The tracking camera is able to estimate the odometry of the robot's floating base which is $[q_{x,y,z,\psi,\theta,\phi},\dot{q}_{x,y,z,\psi,\theta,\phi}]^T$ by Visual Inertial Odometry~(VIO) demonstrated in~\cite{leutenegger2015keyframe} at 30~Hz.
The RGB-Depth camera is utilized to perceive the robot's surroundings by filtered pointclouds and runs at 30~Hz.

\subsection{Mapping}\label{subsec:mapping-details}
A RTAB-Map presented in~\cite{labbe2019rtab} is used in real time to localize the perceived historical pointclouds in the map by optimizing over the 3D map's graph with proximity detection and estimated robot odometry, see~\cite{labbe2018long}. 
The RTAB-Map updates at 1~Hz and the resolution of the voxels is 0.1~m in 3D space. 
In this way, the perceived 3D environment can be described by an Octomap in~\cite{hornung2013octomap} as shown in Fig.~\ref{fig:slam}. 
With an assumption that the robot walks on a flat ground, the obstacles on the ground can be extracted by removing the voxels that are lower than a given threshold. 

\subsubsection{Global Map}
\begin{figure}[t]
    \centering
    \includegraphics[width=\linewidth]{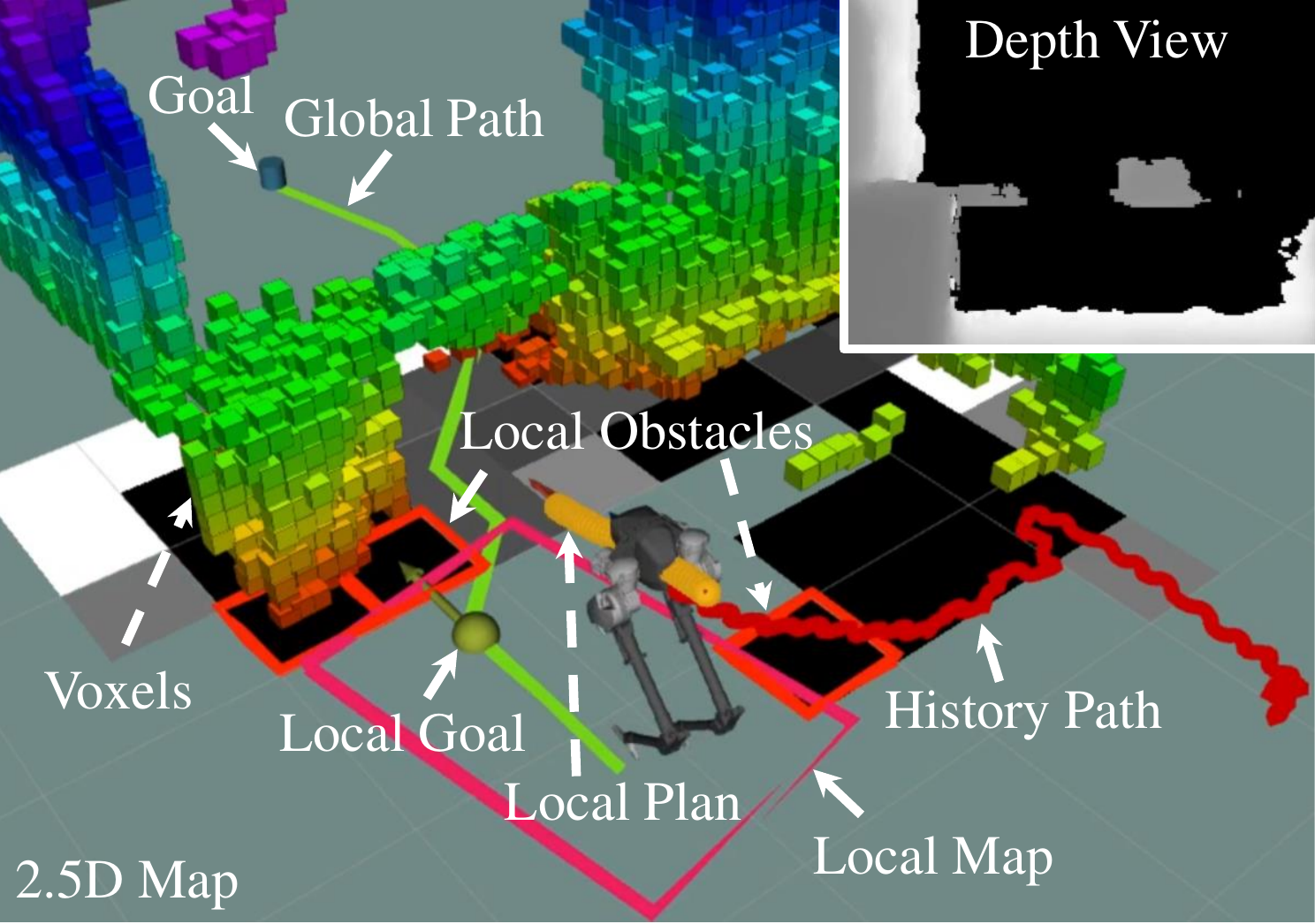}
    \caption{Visualization of localization, mapping, and planning data during navigation. The 3D environment is perceived by the onboard depth camera whose view is shown in the upper right corner and is registered as voxels with a resolution of 0.1~m. A 2.5D map that has 0.5~m resolution is updated accordingly. In this map, regions that are black represent obstacles, white for free space, and admissible height is encoded by gray-scale and the rest is the unexplored region. During planning, the global path shown as a green path is firstly found to lead the robot to reach the goal location represented by the blue cylinder. The local goal that is 1-meter-ahead of the robot is located on the global path and is marked as the green point. A local map drawn as pink bounding box is created with respect to the local goal and robot current position. The obstacles inside the local map is used in the trajectory planners as safety constraints, and they are highlighted by red bounding boxes. The output from the local planner is illustrated by a yellow path. The history of estimated translational position of Cassie's pelvis in 3D is recorded as a red path.}
    \label{fig:slam}
\end{figure}

In order to utilize the spatial information for spontaneous planning, the 3D map is later translated to a 2.5D map where the minimum height of the voxels on the Octomap is recorded on the corresponding 2D cell on the grid, as exemplified in Fig.~\ref{fig:slam}.
The minimum height of the voxels over the ground at the same 2D position is termed as \textit{admissible height} to the robot, because it describes the maximal walking height on that 2D cell. 
\begin{remark}
The sensor stack on top of Cassie adds a 0.25~m vertical dimension to the pelvis height $q_z$, as shown in Fig.~\ref{fig:cassie-model}. In this work, \textit{admissible height} refers to the maximum walking height $q_z$ and \textit{ceiling height} includes this 0.25~m vertical dimension. 
\end{remark}

Compared to Octomap, the 2.5D map has a size of 20~$m^2$ and has a lower resolution of 0.5~m. 
The admissible height recorded in this map is obtained by finding the lowest height of the 0.1~m resolution voxels inside the low-resolution cell, as shown in Fig.~\ref{fig:slam}. 
Such low resolution grid is able to take the robot shape into account because Cassie's footprint radius is around 0.2~m.
In this way, each cell on this 2.5D map can be classified into four types of space for the robot with respect to its admissible height:
\paragraph{Obstacle} The admissible height is lower than the robot's lowest walking height (0.7~m) and is, therefore, not traversable for the robot.
\paragraph{Height-constrained} The admissible height is within the robot's lowest and largest walking height~(1~m) because such a region is traversable but requires the robot to lower its walking height to be below this admissible height.
\paragraph{Free} The robot can travel through it using its largest walking height. 
\paragraph{Unexplored} The low-resolution cell that has not been well perceived and does not have obstacles, \textit{i.e.}, only part of this cell is detected as free space.

\begin{figure*}[t]
    \centering
    \begin{subfigure}{0.275\textwidth}
        \centering
        \includegraphics[width=\linewidth]{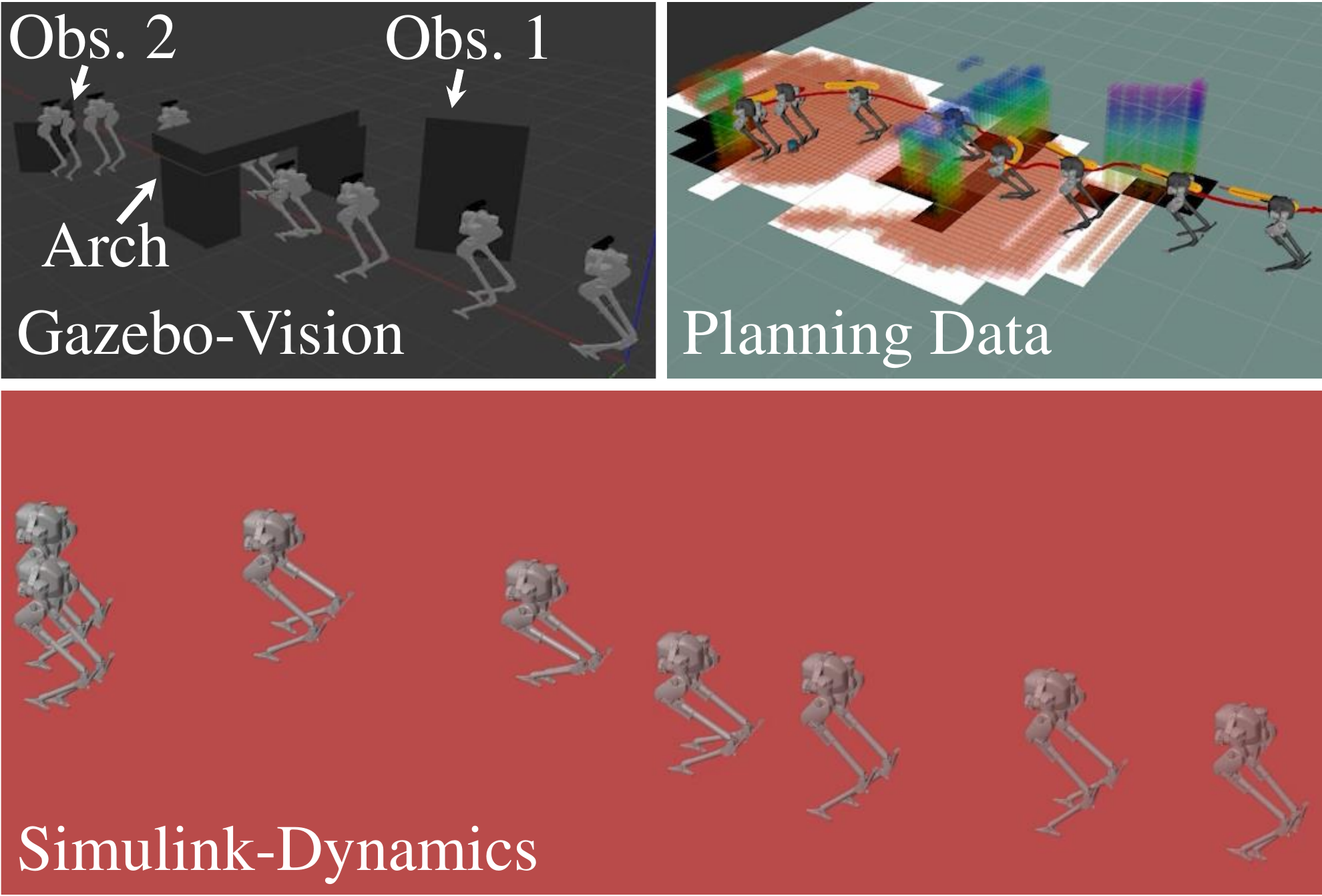}
        \caption{A snapshot of joint simulation of vision and dynamics using the proposed method in a congested space} \label{subfig:sim_baseline_vis}
    \end{subfigure} 
    \begin{subfigure}{0.215\textwidth}
        \centering
        \includegraphics[width=\linewidth]{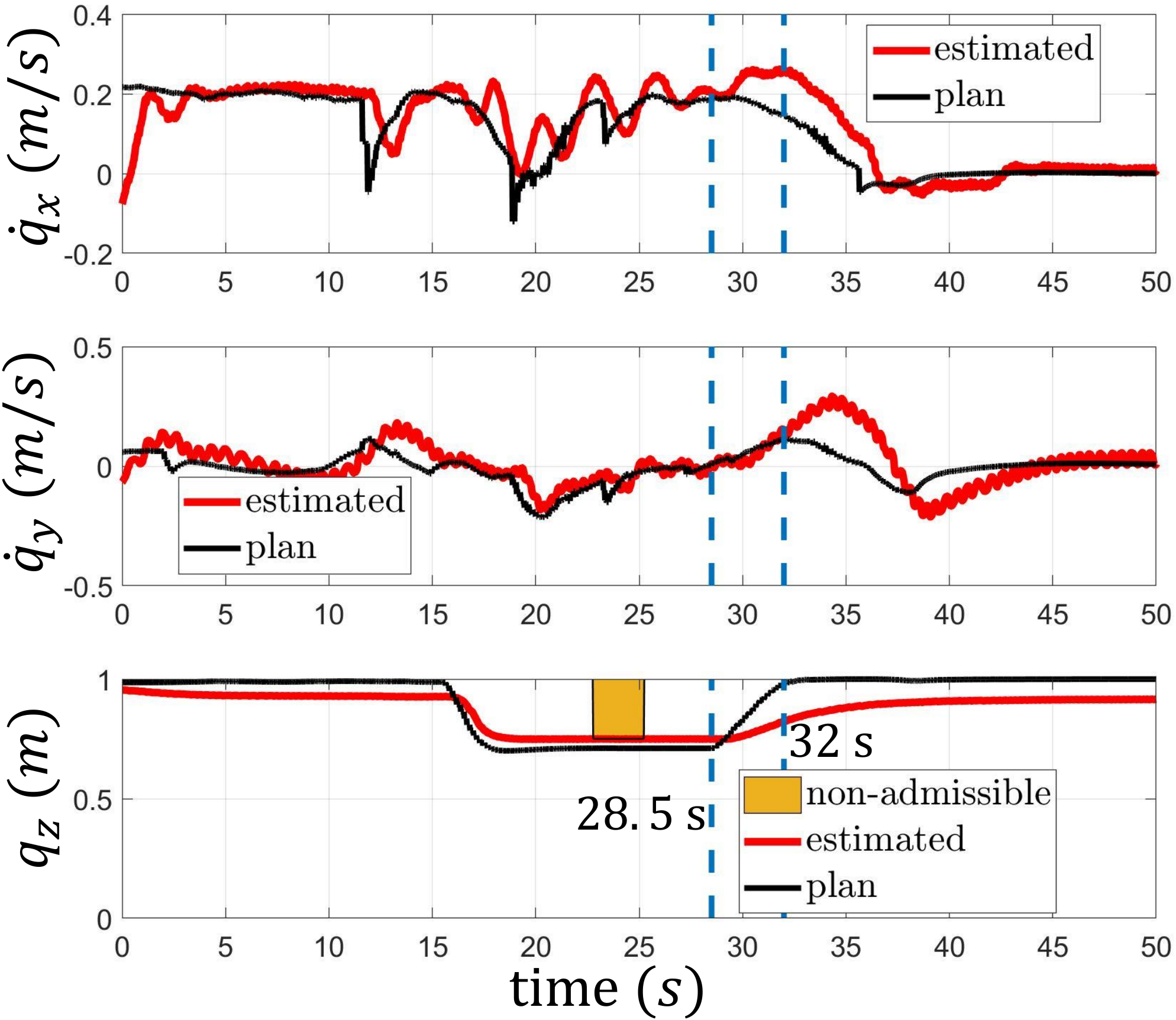}
        \caption{Planned and actual walking speeds and walking height using the proposed method} \label{subfig:sim_baseline_plot}
    \end{subfigure}
    \begin{subfigure}{0.275\textwidth}
        \centering
        \includegraphics[width=\linewidth]{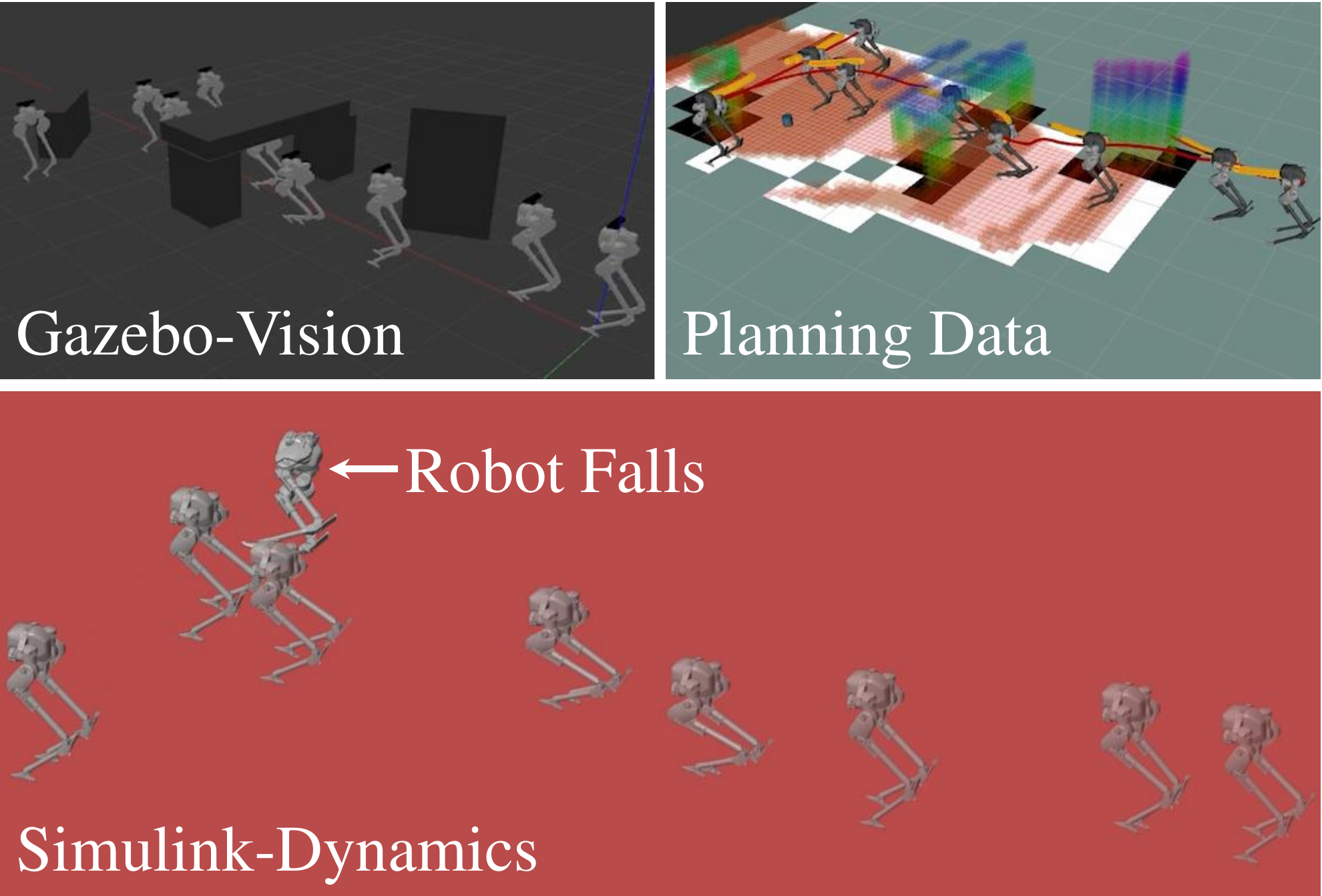}
        \caption{A snapshot of joint simulation of vision and dynamics without feasible command set in the same scenario} \label{subfig:sim_nofeas_vis}
    \end{subfigure} 
    \begin{subfigure}{0.215\textwidth}
        \centering
        \includegraphics[width=\linewidth]{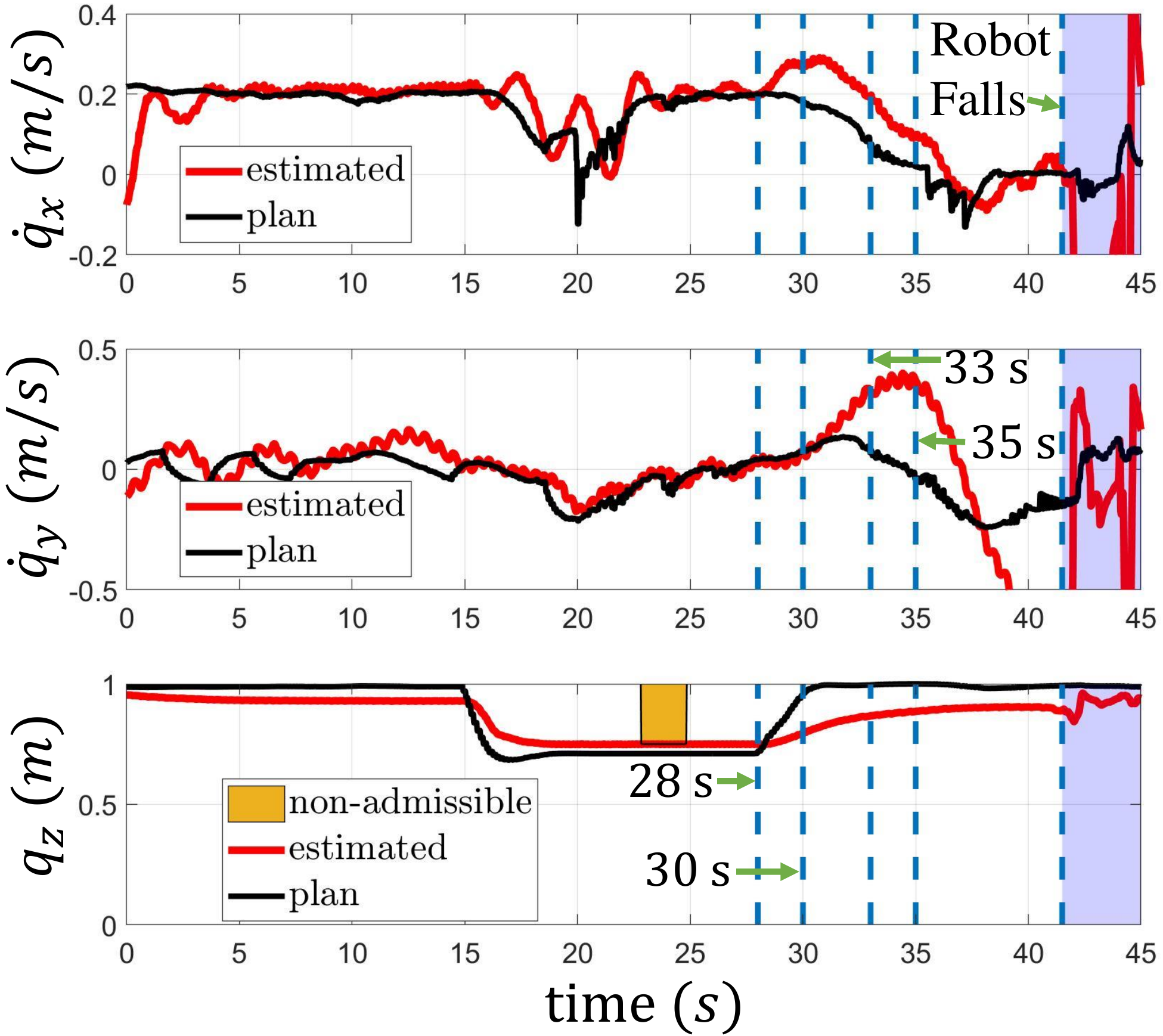}
        \caption{Planned and actual walking speeds and walking height without feasible command set} \label{subfig:sim_nofeas_plot}
    \end{subfigure}    
    \caption{Validation of the proposed navigation autonomy in the joint simulation. A virtual congested space is built in Gazebo where the robot's depth camera reading is simulated while the robot's dynamics is synchronously computed in MATLAB Simulink. In Fig.~\ref{subfig:sim_baseline_vis}, the proposed navigation framework enables Cassie to successfully avoid the first obstacle (Obs.~1), lower the walking height to travel underneath an arch, recover to the normal height and stop at the goal location while avoiding collision with the second obstacle (Obs.~2). However, the autonomy which does not consider the feasible command set introduced in Sec.~\ref{subsec:safeset} causes the robot to fall over in Simulink, as shown in Fig.~\ref{subfig:sim_nofeas_vis}.}    
    \label{fig:sim_baseline}
\end{figure*}

In this way, planners can utilize the information on this 2.5D map to find an optimal path to navigate the robot while considering the varying admissible height and obstacles in the environment. 
The global planner is deployed on this 2.5D global map while trajectory planners developed in Sec.~\ref{sec:local-planners} only use the information of a small region on the global map which are denoted as a local map.

\subsubsection{Local Map}
For optimization-based trajectory planners that only plan for short trajectories, the information over the entire map is not necessary. 
Therefore, a local map that only contains the environment information of the robot's surroundings is built. 
In this work, the local map is designed to be a bounding box whose width is 1.2~m and length is 2.75~m, and is shown in Fig.~\ref{fig:slam} as the pink bounding box. 
The location and orientation of this local region can be well defined by the robot's current position and the local goal for the local planner and can be updated in real time.
The locations of all the obstacles, \textit{i.e.}, 0.5~m untraversable cells, inside this local map will be considered in the collision-free constraints via~\eqref{eq:safeset_collision_free} in Sec.~\ref{sec:local-planners}, which are termed as local obstacles in Fig.~\ref{fig:slam}. The minimum admissible height $\underline{h}$ in this region will be added to the height constraint by~\eqref{eq:height_constraints}.

\subsection{Global Planner}\label{subsec:global_planner}
\subsubsection{A* Search}
In the scenario where the robot is exploring an unknown environment, a global planner is needed to quickly find a collision-free path (a list of waypoints) to lead the robot to reach the goal location on the global map and should be able to replan when the map is updated.
Moreover, in the height-constrained environment, we prefer the robot to avoid traveling underneath obstacles if the robot can walk around the obstacles.
Therefore, an A* search presented in~\cite{hart1968formal} that encodes the costs of different types of regions on the map is introduced to serve as the global planner. 
The cost $\chi(n)$ of each node $n$, \textit{i.e.}, traversal cell, during the search is defined as $\chi(n)=\gamma(n)+\eta(n)$ where $\gamma(n)$ is the cost of coming from the start node to the current node while $\eta(n)$ is the heuristic cost that is defined as 
\begin{equation}
    \eta(n)=
    [w_d, w_h]
    \begin{bmatrix}
    \eta_d(n) \\ \eta_h(n) \\
    \end{bmatrix},
    \begin{cases} [w_d, w_h]=[1,3],&\text{explored}  \\ 
    [w_d, w_h]=[1.2,0],&\text{otherwise} \end{cases} 
\end{equation}
where $\eta_d(n)$ is the Euclidean \textit{distance} to the goal location and $\eta_h(n)$ is the cost of admissible \textit{height} of the node, \textit{i.e.}, a node that has lower admissible height will have a larger $\eta_h(n)$. 
In this way, the A* will prefer to include the node which has a larger admissible height into the path. 
Moreover, a node that has not been explored will be avoided if there is a well-known node around, and this is achieved by having a larger weight $w_d$ on the node that is unexplored.  
In this way, the A* search is able to find an optimal path that is shortest to the goal location while avoiding height-constrained and unexplored regions if possible.

\subsubsection{Local goal}
This global planner returns a sequence of waypoints which are the 2D coordinates of the selected nodes in the planned path and these are updated at 1~Hz.   
The local goal that is ahead of the robot's current position in a given range will be selected from this global path by a breadth-first search, as marked in Fig.~\ref{fig:slam}. 
In this work, the local goal is chosen as 1-meter ahead of the current estimated robot position. 
Specifically, the nearest two waypoints around one-meter ahead are found first and then a linear interpolation between these two waypoints allows us to select the local goal.
This local goal selected by the proposed approach is continuous with respect to the robot's movement and is passed to the local planner as part of the target state via~\eqref{eq:final_constraint} in Sec.~\ref{sec:local-planners}.

By now, the infrastructure for the autonomy is developed, and the trajectory planners and walking controller can be embedded in this framework to drive the bipedal robot to safely explore height-constrained environments as illustrated in Fig.~\ref{fig:framework}. 
The entire pipeline is validated next in simulations and experiments in the following sections.
\section{Numerical Validation} \label{sec:simulation}
In order to validate the proposed navigation framework and exemplify the necessity of the feasible command set introduced in Sec.~\ref{subsec:safeset} and used in Sec.~\ref{sec:local-planners}, the proposed navigation autonomy is deployed on the bipedal robot Cassie and is tested extensively in a joint Simulink-Gazebo simulation for simulating dynamics and vision.

\subsection{Simulation Implementation}~\label{subsec:sim_implementation}
We utilize Gazebo to simulate the robot's depth camera reading and MATLAB Simulink to calculate the robot full-order dynamics in order to have a higher fidelity simulation of the robot's walking dynamics. 
These two simulators are synchronized and the navigation algorithms are running online during simulation. Specifically, the simulation of robot dynamics and vision will not wait for the results of planners and the controller, which is aligned with the real-world setting. 
The planning data is visualized via ROS RViz, as shown in Fig.~\ref{subfig:sim_baseline_vis},\ref{subfig:sim_nofeas_vis}.

For the simulation test, a congested space with an arch whose admissible height is 0.75m and two obstacles are constructed in Gazebo, as demonstrated in Fig.~\ref{fig:sim_baseline}. 
The initial position of the robot is behind the first obstacle and the goal location is set to be close to the second obstacle.

\subsection{Validation of the Proposed Autonomy}
The proposed navigation pipeline shown in Fig.~\ref{fig:framework} is first tested in the simulation. It demonstrates the capacity to drive the robot to avoid the first obstacle, crouch down to pass through the arch, recover to a normal walking height afterward, and then stop in front of the second obstacle. All this is done without a single fall or collision, as exhibited in Fig.~\ref{subfig:sim_baseline_vis}. The profiles of the robot's planned and actual walking speeds and walking height are recorded in Fig.~\ref{subfig:sim_baseline_plot}.

\subsection{Autonomy without Feasible Command Set}
In order to demonstrate the necessity of the feasible command set during the navigation, we consider the pipeline as introduced in Sec.~\ref{sec:local-planners} but remove the feasible command set which introduces a gait stability constraint via~\eqref{eq:safeset}. This is compared with the original pipeline without changes.  The parameters used in the walking controller, cost functions~\eqref{eq:cost-function}, and the rest of the constraints in Sec.~\ref{subsec:constraints} are kept identical to obtain a fair comparison.

As shown in Fig.~\ref{subfig:sim_nofeas_vis}, the navigation autonomy without the feasible set causes the robot to fall. This occurs when the robot is approaching the goal location while increasing the walking height after the robot walks through the arch and avoids collision with the second obstacle.

According to the recorded data in Fig.~\ref{subfig:sim_nofeas_plot}, during the period of 28~s to 30~s, when the robot is changing its walking height from 0.75~m to 1~m while walking to the left, the profile of planned walking height is much steeper than the one using~\eqref{eq:safeset} (28.5~s to 32~s in Fig.~\ref{subfig:sim_baseline_plot}). 
Because of the coupled dynamics between planar walking velocity and walking height, such fast changes in the walking height triggers a larger robot lateral walking speed under similar planar walking velocity commands. 
As a result, the robot's \textit{actual} lateral speed accelerates to over 0.4~m/s during 33~s to 35~s, and the planner needs to use a large lateral speed command later during 36~s to 40~s to drive the robot back.
At this period, the lateral walking speed command is over -0.2~m/s at $q_z$ of 1~m which lies outside of the feasible command set in Fig.~\ref{fig:safe_command_set}, and directly results in gait instability.

This showcases the importance of the feasible command set.
Without it, the planned trajectory is not \textit{dynamically feasible} in the sense that the plan does not take into account dynamic coupling that exists between the planar velocity and the walking height, leading to unbounded robot states and eventually a walking failure like Fig.~\ref{subfig:sim_nofeas_vis},\ref{subfig:sim_nofeas_plot}.

We repeated such ablation test at least 5 times with different robot initial states in this simulation and results are consistent across all trials.
More extensive tests on the proposed autonomy are conducted in the experiments on the real robot Cassie in the following section.

\section{Experiments} \label{sec:experiments}

\begin{figure*}[!htp]
    \centering
    \begin{subfigure}{0.265\textwidth}
        \centering
        \includegraphics[width=\linewidth]{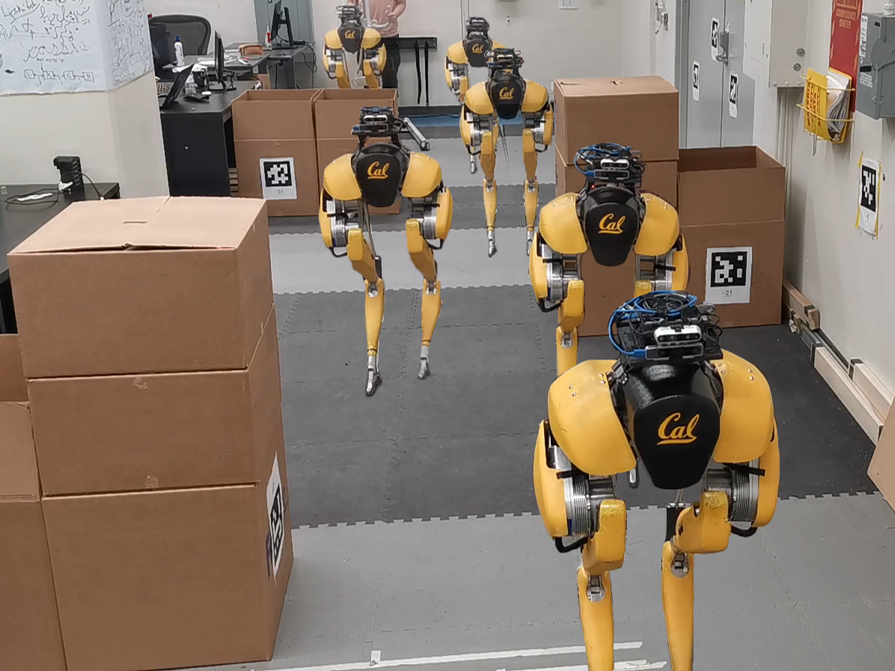}
        \caption{A snapshot of experiment in the 2D maze} \label{subfig:maze_expr}
    \end{subfigure} 
    \begin{subfigure}{0.235\textwidth}
        \centering
        \includegraphics[width=\linewidth]{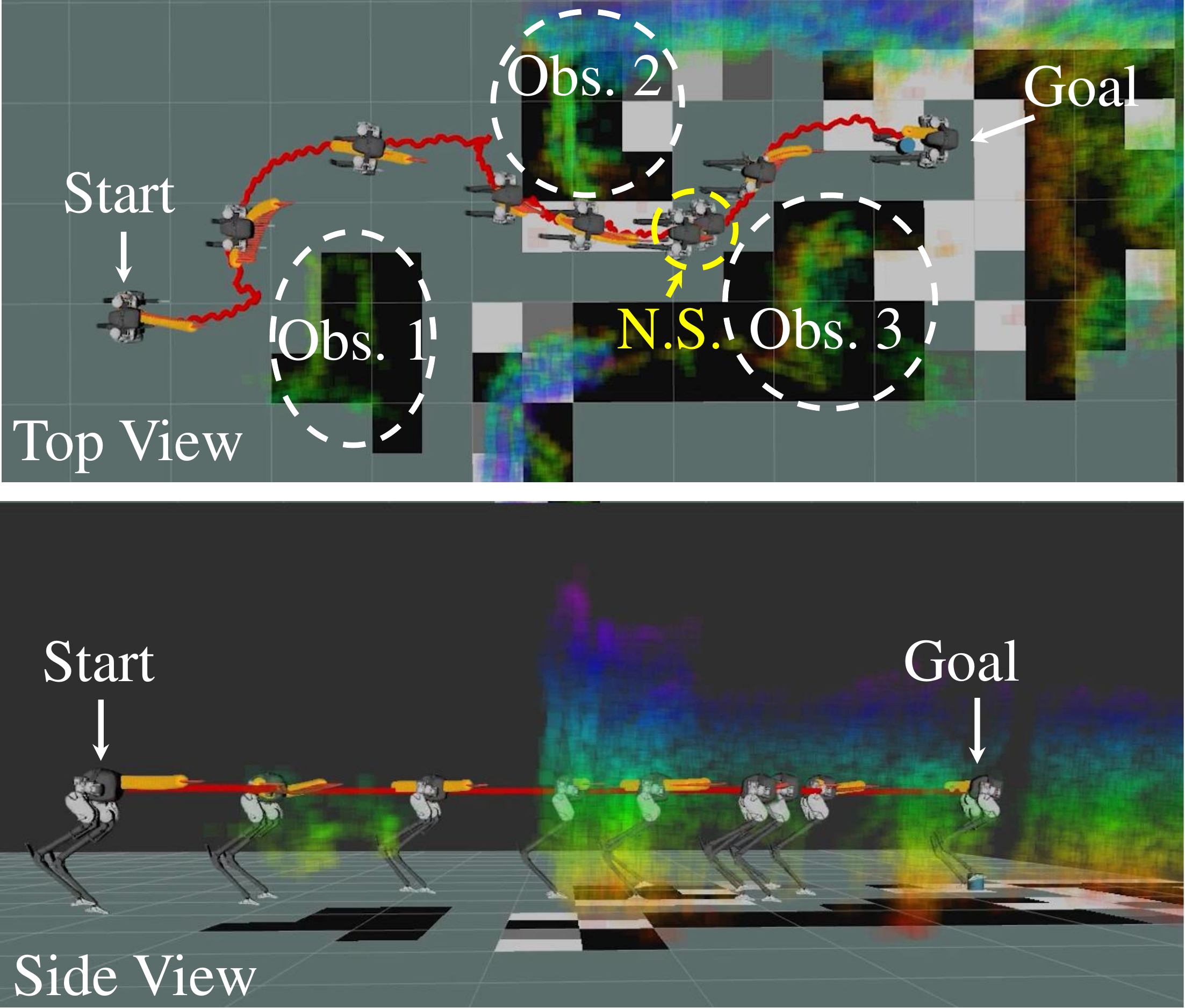}
        \caption{Planning snapshot during the experiment in the 2D maze} \label{subfig:maze_rviz}
    \end{subfigure} 
    \begin{subfigure}{0.24\textwidth}
        \centering
        \includegraphics[width=\linewidth]{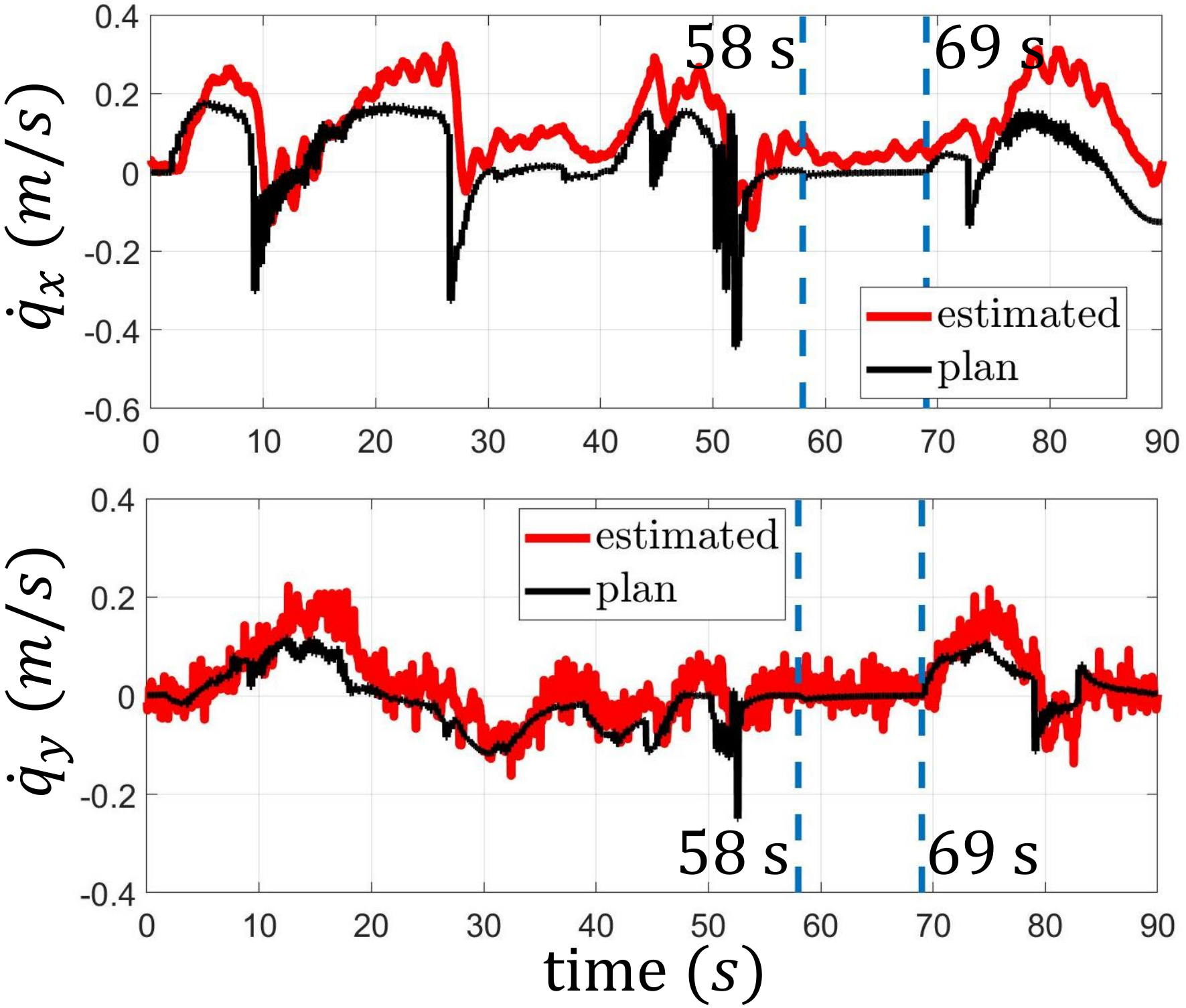}
        \caption{A comparison b/w planned and actual walking speeds} \label{subfig:maze_vxvy}
    \end{subfigure}
    \begin{subfigure}{0.24\textwidth}
        \centering
        \includegraphics[width=\linewidth]{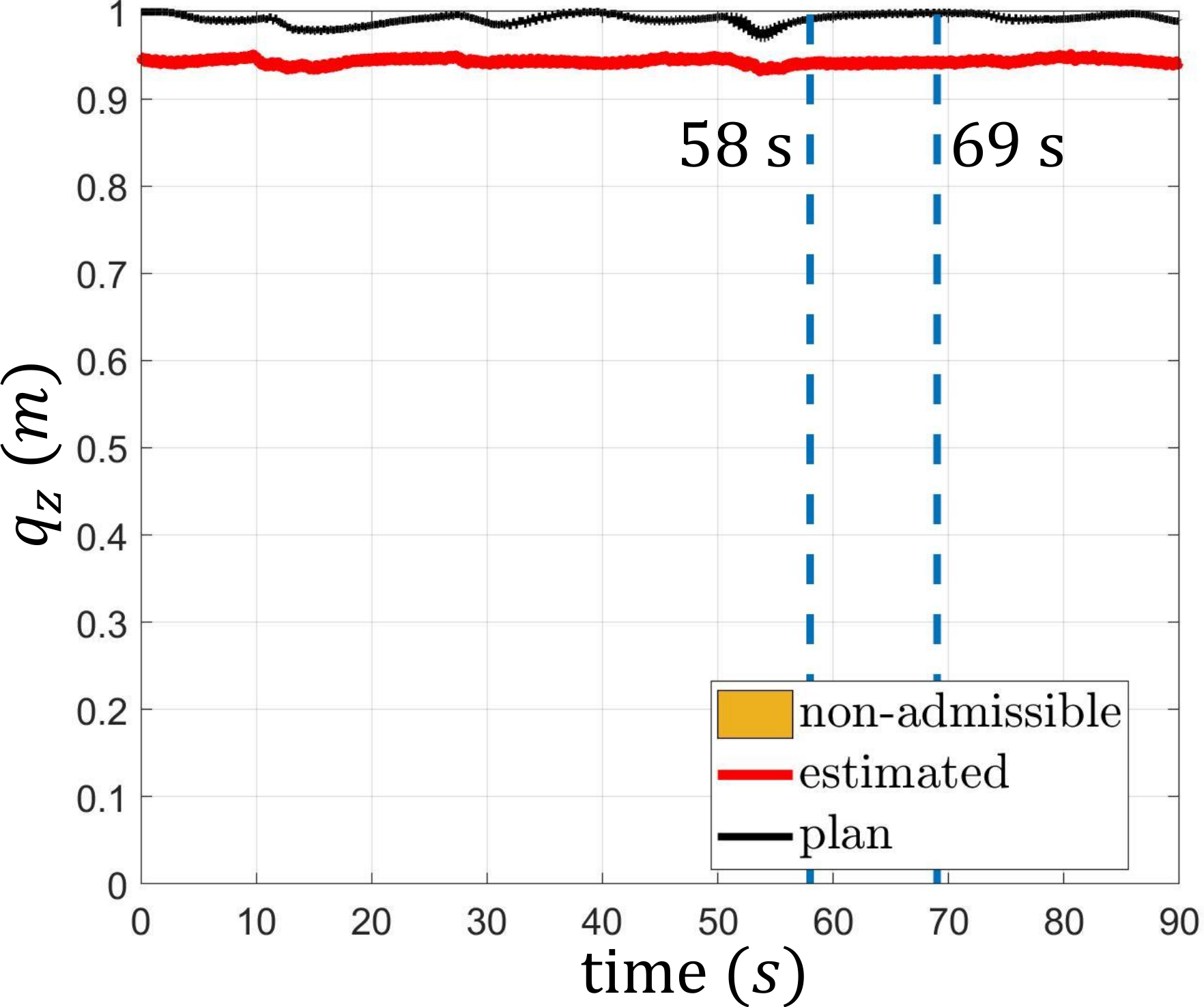}
        \caption{A comparison b/w planned and actual walking height} \label{subfig:maze_wh}
    \end{subfigure}
    \caption{2D Maze Experiment. In this scenario, Cassie walks through a narrow space where three Obstacles (Obs.~1-3) are placed in the 4~m$\times$10~m area and are formed as a maze for the robot without height constraint. Without prior information on the global map, Cassie shows the ability to avoid the obstacle that is updated in real time while traveling to the goal location. During the experiment, there is a Narrow Space~(N.S.) circled by the yellow dash line in Fig.~\ref{subfig:maze_rviz} that is only 0.5~m$\times$0.5~m in shape. Cassie slows down and uses minimal walking velocity in this region, as shown from 58~s to 69~s in Fig.~\ref{subfig:maze_vxvy}. The autonomy framework keeps the robot walking after it finds a safe path to get out of such a narrow area after 69~s. Moreover, the reactive local planner commands several sudden jumps of the planned walking speeds to negative in Fig.~\ref{subfig:maze_vxvy} to ensure the robot's safety when the robot is close to the obstacles.}
    \label{fig:maze_result}
\end{figure*}

\begin{figure*}[!htp]
    \centering
    \begin{subfigure}{0.265\textwidth}
        \centering
        \includegraphics[width=\linewidth]{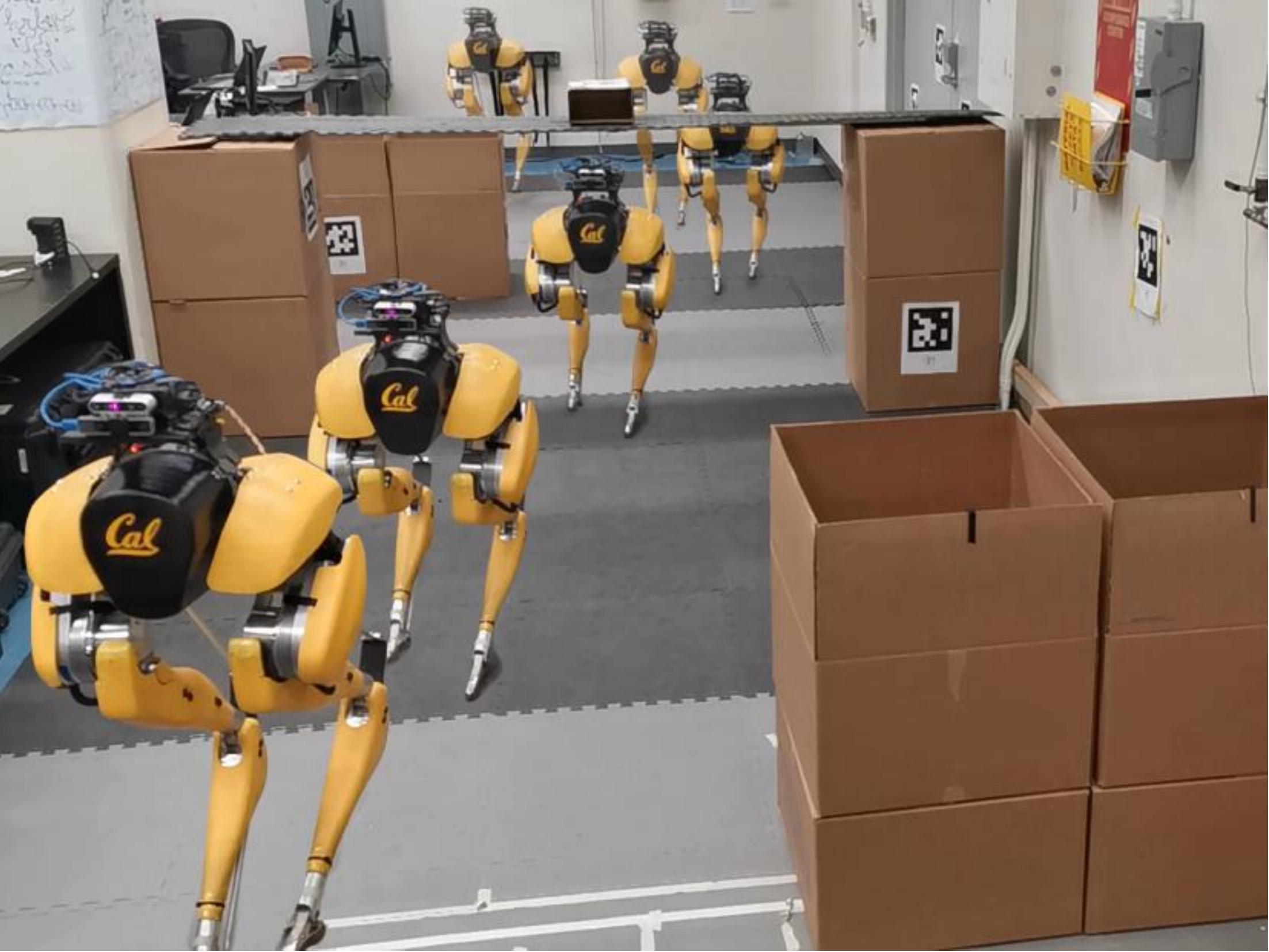}
        \caption{A snapshot of experiment in the height-constrained space} \label{subfig:clutter_expr}
    \end{subfigure} 
    \begin{subfigure}{0.235\textwidth}
        \centering
        \includegraphics[width=\linewidth]{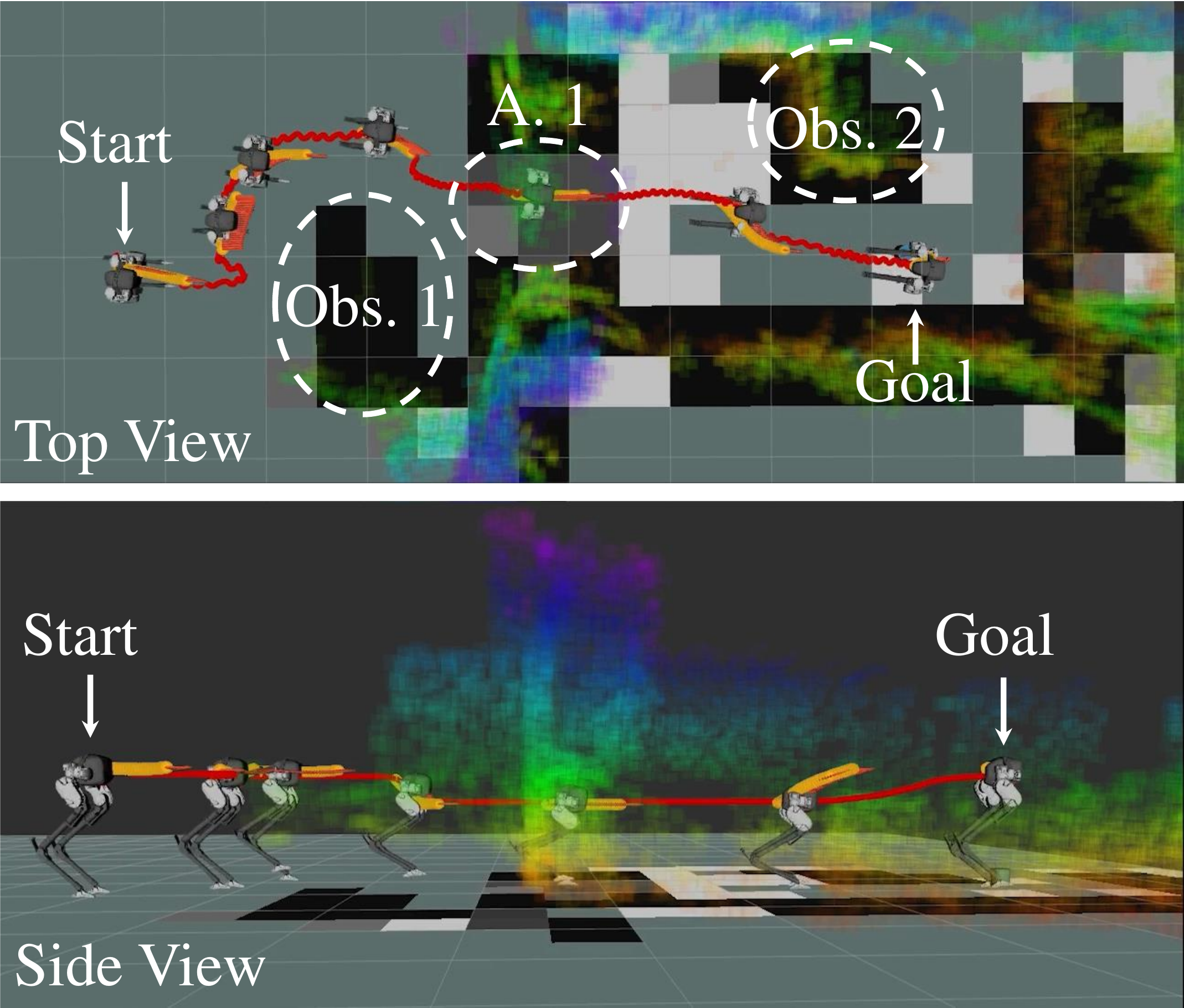}
        \caption{Planning snapshot during the experiment} \label{subfig:clutter_rviz}
    \end{subfigure} 
    \begin{subfigure}{0.23\textwidth}
        \centering
        \includegraphics[width=\linewidth]{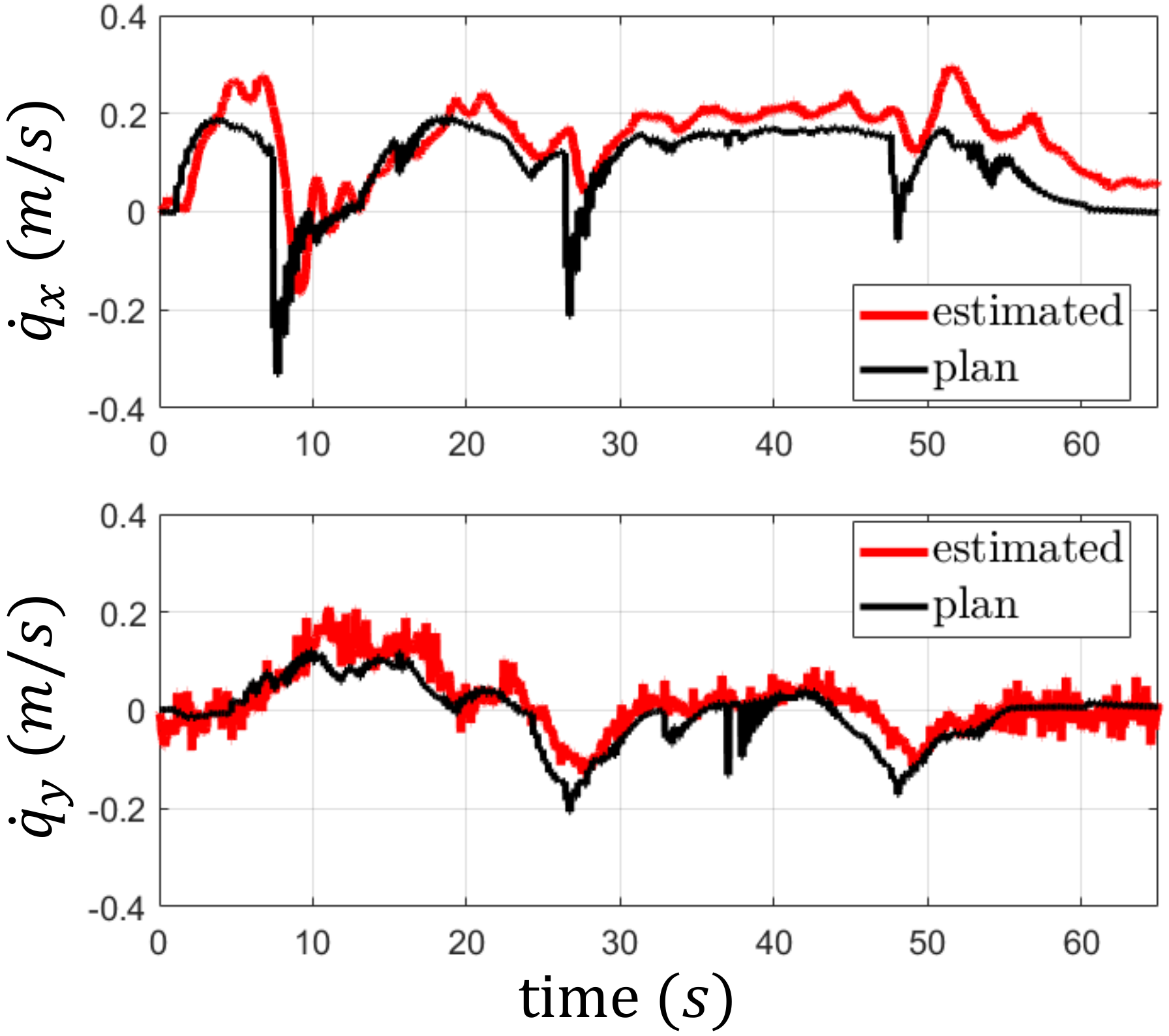}
        \caption{A comparison b/w planned and actual walking speeds} \label{subfig:clutter_vxvy}
    \end{subfigure}
    \begin{subfigure}{0.245\textwidth}
        \centering
        \includegraphics[width=\linewidth]{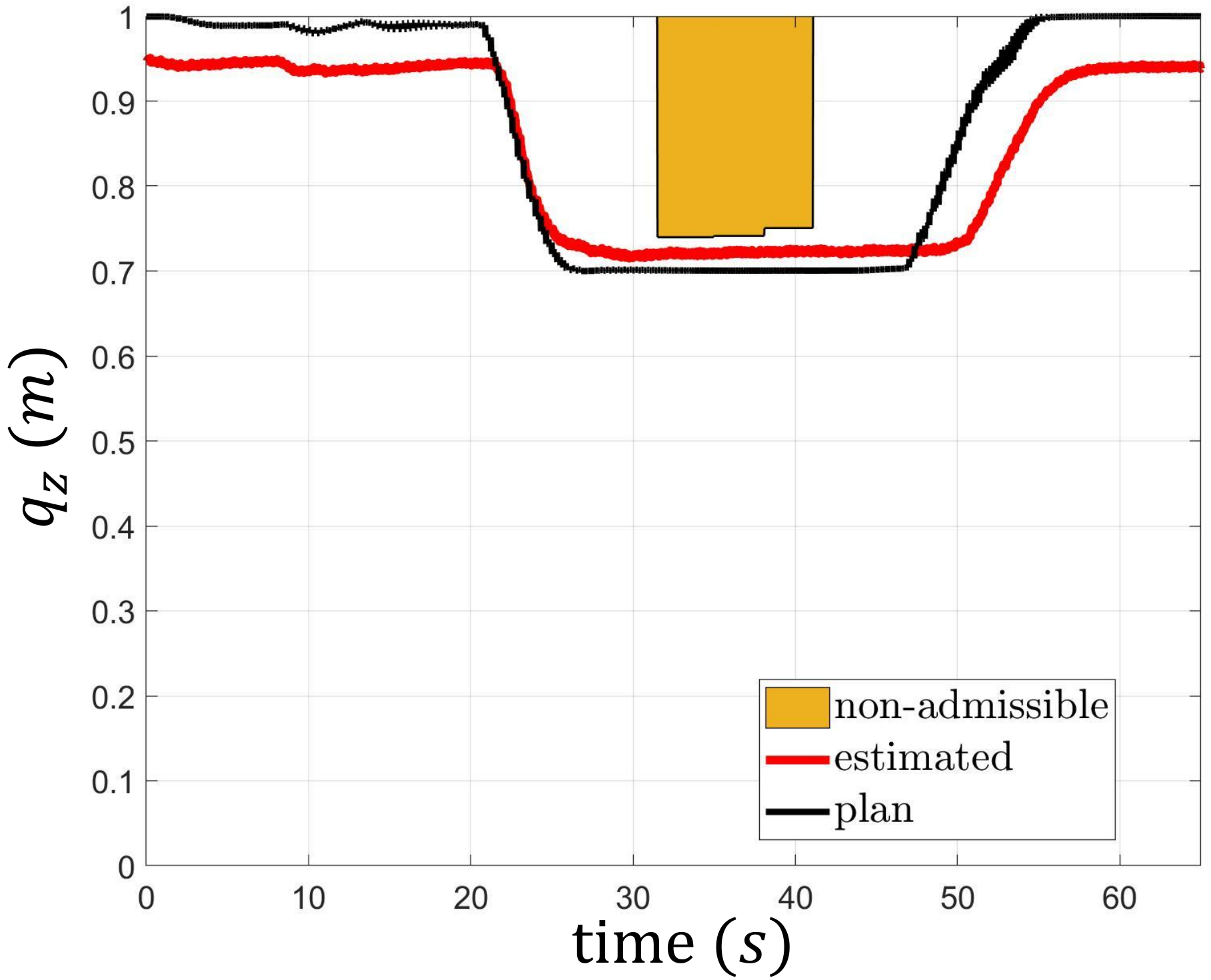}
        \caption{Planned, actual and non-admissible walking height} \label{subfig:clutter_wh}
    \end{subfigure}    
    \caption{Height-Constrained Space Experiment. There are two Obstacles~(Obs.~1-2) and one Arch~(A.~1) area as circled in Fig.~\ref{subfig:clutter_rviz} in this scenario. During the exploration, Cassie firstly avoids Obs.~1, then crouches down to walk underneath the overhanging obstacle in Arch A.~1. In this area, the overhanging obstacle has only 1~m clearance to the ground, 1~m in width, and 1.5~m in length, and the robot needs to use a walking height below 0.75~m to avoid collision, as recorded in Fig.~\ref{subfig:clutter_wh}. After passing this region, Obs.~2 appears in the robot's view and Cassie quickly reroutes to avoid it until it reaches the goal location.}
    \label{fig:clutter_result}
    \vspace{-0.2cm}
\end{figure*}

\subsection{Hardware Implementation}
Cassie has a real-time computer running the variable walking height controller and kinematics-based state estimation at 2~kHz.
There are two Intel NUC mini computers with i7 processors used for the entire navigation pipeline.
One computer processes the perceived spatial information and registers it into the 2.5D map. 
It also communicates with the Cassie's computer in order to send real-time commands to the controller while processing robot's state feedback. 
Another mini computer receives the maps and deals with all planning-related work, data visualization, and provides the planned control commands.
The testing cases are indoor scenarios where obstacles are represented by cardboard boxes. 
Note that while there are some QR codes used in the environments, like the ones in Fig.~\ref{subfig:maze_expr}, these are only used to add more features to a relative-plain background to enhance the performance of VIO and are not exploited to store prior information of ground truth locations in the environment.

\subsection{Experiments}
In this work, the autonomy is tested in four types of indoor environments which include (1) a maze without height constraints, (2) a cluttered space with multiple obstacles and one height-constrained space in the form of an arch, (3) an area that has multiple arches and a ground obstacle, and (4) an obstructed door. 
The motivation of this study is to validate the proposed legged navigation autonomy that can handle unknown environments with a range of obstacles. 
Each experiment is designed to test a specific aspect of the algorithm's performance and build upon the previous experiment.

The proposed autonomy is tested successfully in at least 4 trials where the robot is able to safely walk to the given goal location without a single fall in each of these scenarios. 
Representative experiments (video\footnote{\url{https://youtu.be/Da0tebC3WuE}}) are exemplified and analyzed below.

\subsubsection{2D Maze}
In this scenario, three obstacles are distributed in a cluttered space with 4~m width and 10~m length without any height constraints, as shown in Fig.~\ref{subfig:maze_expr}.
Each of the obstacles is in the shape of 0.5~m$\times$1.0~m and they form a maze for the robot. 
In order to test the ability of the autonomy to explore an unknown and complex environment, Cassie is initialized in front of the first obstacle which blocks most of the view of the robot and therefore the robot has a limited prior information about the entire maze. 
During the test, as illustrated in Fig.~\ref{subfig:maze_expr},\ref{subfig:maze_rviz}, the global planner is able to quickly reroute and create an updated path to avoid the new detected obstacle, while the local planners update in real-time to regulate the robot movement to avoid collision.
Moreover, several decelerating maneuvers take place when the robot walks close to the obstacles to ensure the robot's safety.
This is realized by outputting a negative walking speed from the reactive local planner as presented in Fig.~\ref{subfig:maze_vxvy}.
Between 58~s to 69~s, Cassie enters a very narrow space whose size is only 0.5~m$\times$0.5~m, a similar size as the robot footprint, as marked in Fig.~\ref{subfig:maze_rviz}. 
At this time, the planners slowly change the robot's movement using almost zero velocity commands, as shown in Fig.~\ref{subfig:maze_vxvy}, in order to keep the robot safe. 
After the robot moves to a more open space, the planners quickly find a trajectory to lead the robot to get out of the current narrow space, which happens after 69~s in Fig.~\ref{subfig:maze_vxvy}. 
As this environment doesn't have any height constraints, the robot maintains a normal height of around 1~m all the time, as exhibited in Fig.~\ref{subfig:maze_wh}.
This experiment demonstrates the capability of the proposed autonomy of a large-scale humanoid robot to safely navigate in a tight space that has almost the same width as the robot's footprint.

\begin{figure*}[!htp]
    \centering
    \begin{subfigure}{0.265\textwidth}
        \centering
        \includegraphics[width=\linewidth]{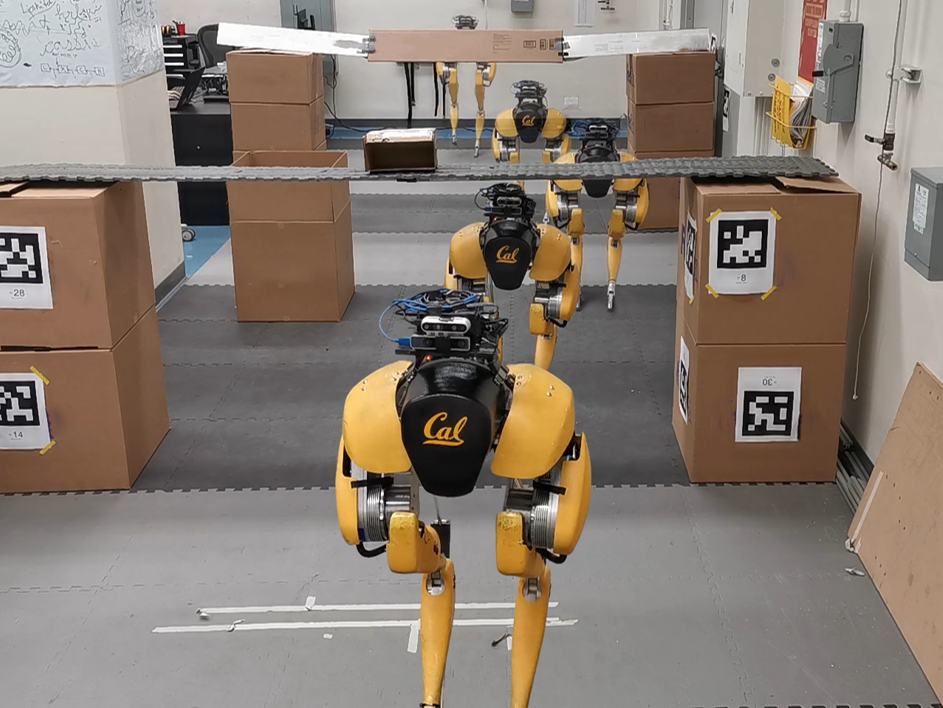}
        \caption{A snapshot of experiment with multiple arches} \label{subfig:tunnel_expr}
    \end{subfigure} 
    \begin{subfigure}{0.235\textwidth}
        \centering
        \includegraphics[width=\linewidth]{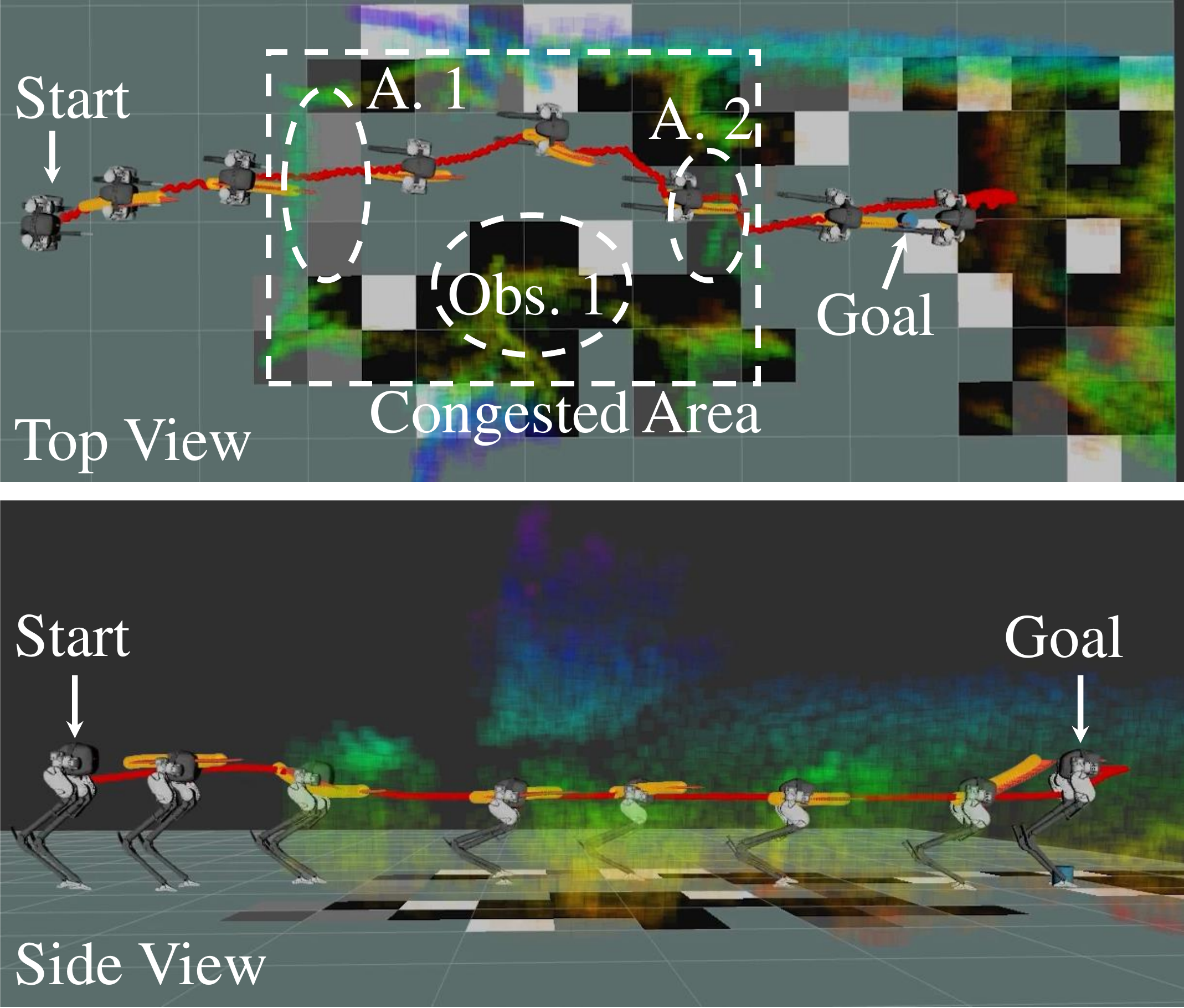}
        \caption{Planning snapshot during the experiment with multiple arches} \label{subfig:tunnel_rviz}
    \end{subfigure} 
    \begin{subfigure}{0.23\textwidth}
        \centering
        \includegraphics[width=\linewidth]{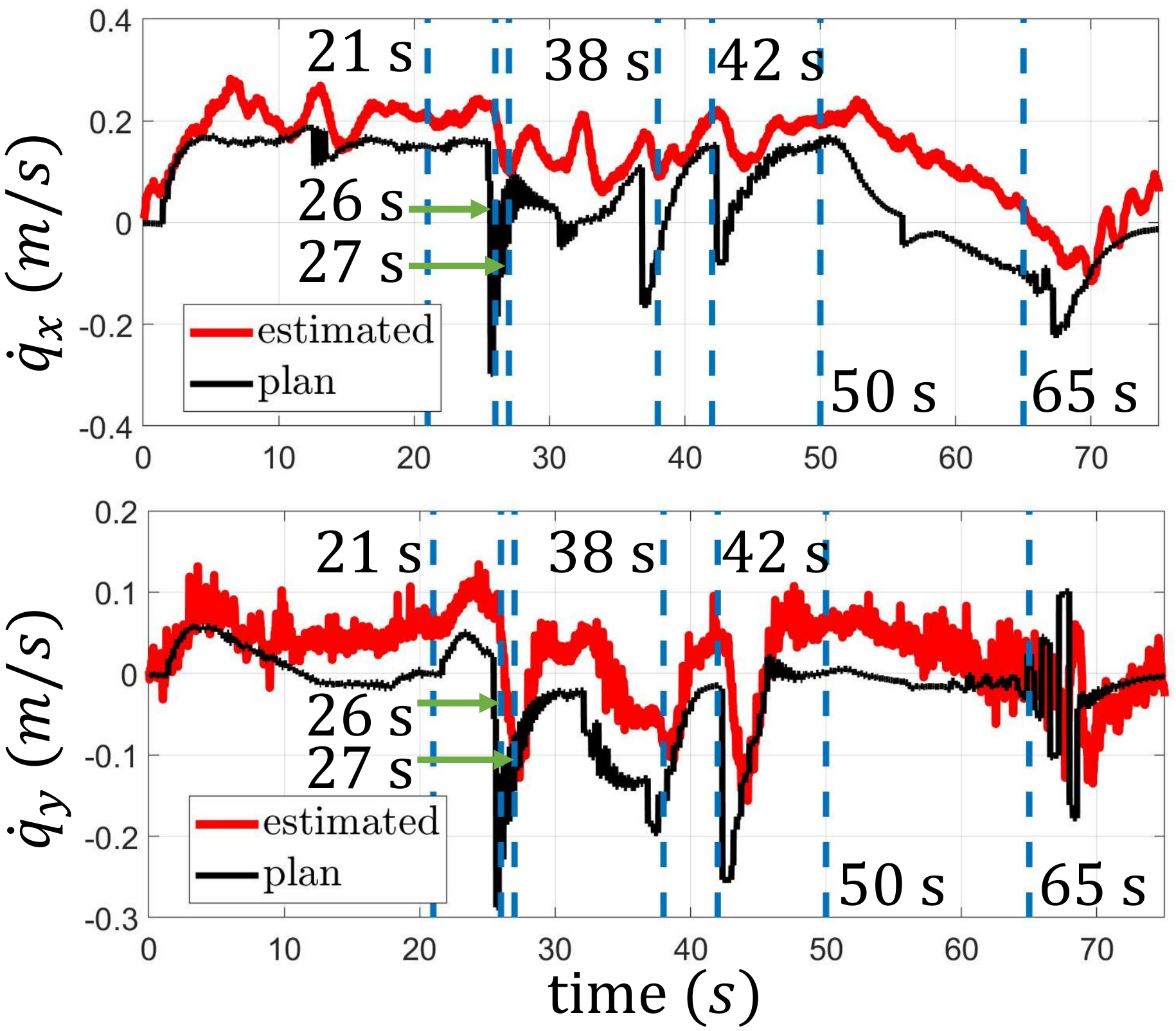}
        \caption{A comparison b/w planned and actual walking speeds} \label{subfig:tunnel_vxvy}
    \end{subfigure}
    \begin{subfigure}{0.245\textwidth}
        \centering
        \includegraphics[width=\linewidth]{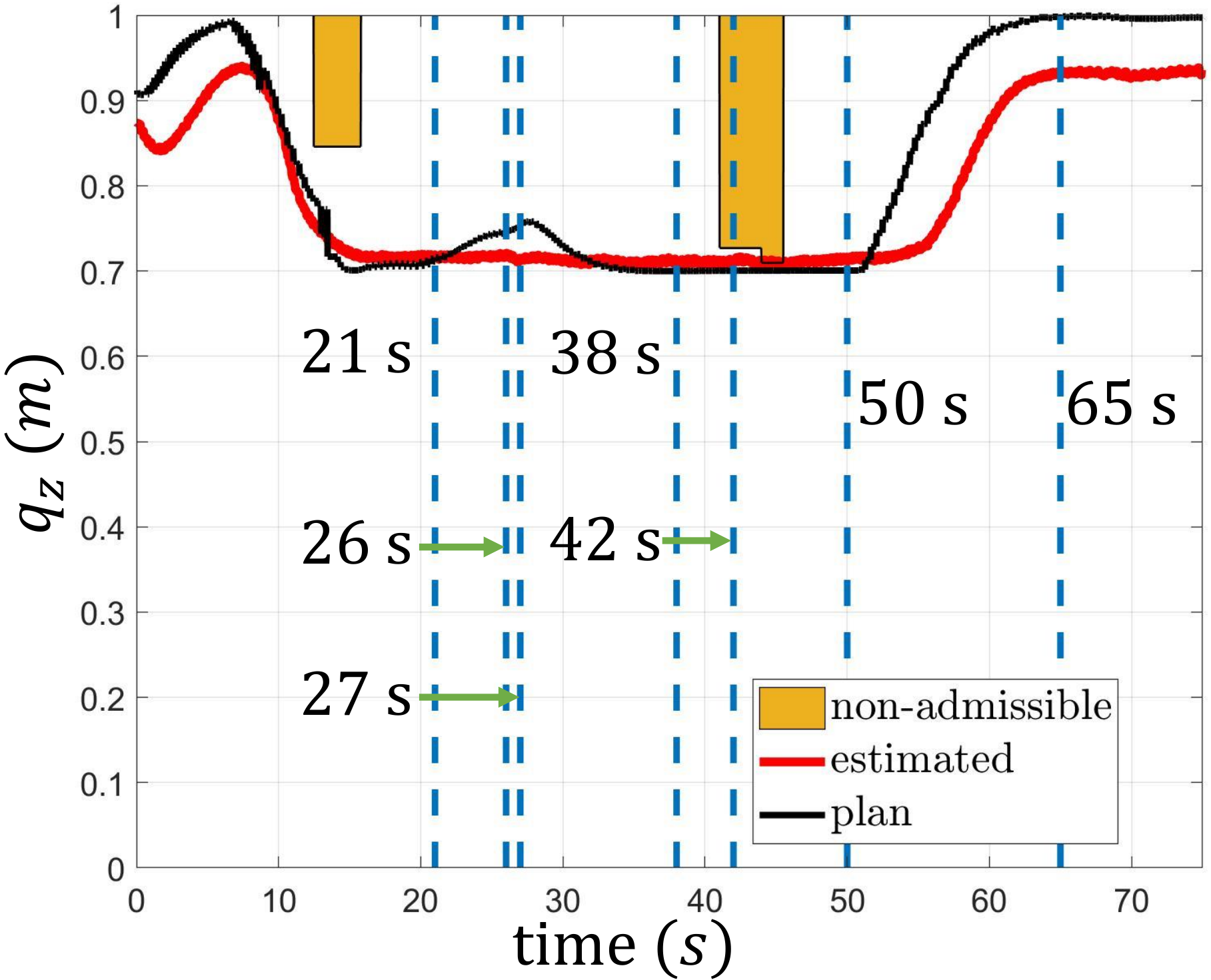}
        \caption{Planned, actual and non-admissible walking height} \label{subfig:tunnel_wh}
    \end{subfigure}    
    \caption{Multiple Arches Experiment. There are two arches~(A.~1, A.~2) placed in the two ends of the 4~m$\times$10~m space and there is one Obstacle~(Obs.~1) between these A.~1 and A.~2, as marked in Fig.~\ref{subfig:tunnel_rviz}. These two height-constrained areas stand for the entrance and exit of a congested area. The admissible height in A.~1 and A.~2 are 0.85~m and 0.75~m, respectively, as illustrated in Fig.~\ref{subfig:tunnel_wh}. There exists sensor noise which causes a 0.71~m admissible height for the second arch recorded in Fig.~\ref{subfig:tunnel_wh}. Still, Cassie is capable to keep a low walking height at 0.7~m to safely travel in this region while avoiding an obstacle (Obs.~1) inside.}
    \label{fig:tunnel_result}
\end{figure*}

\begin{figure*}[!htp]
    \centering
    \begin{subfigure}{0.265\textwidth}
        \centering
        \includegraphics[width=\linewidth]{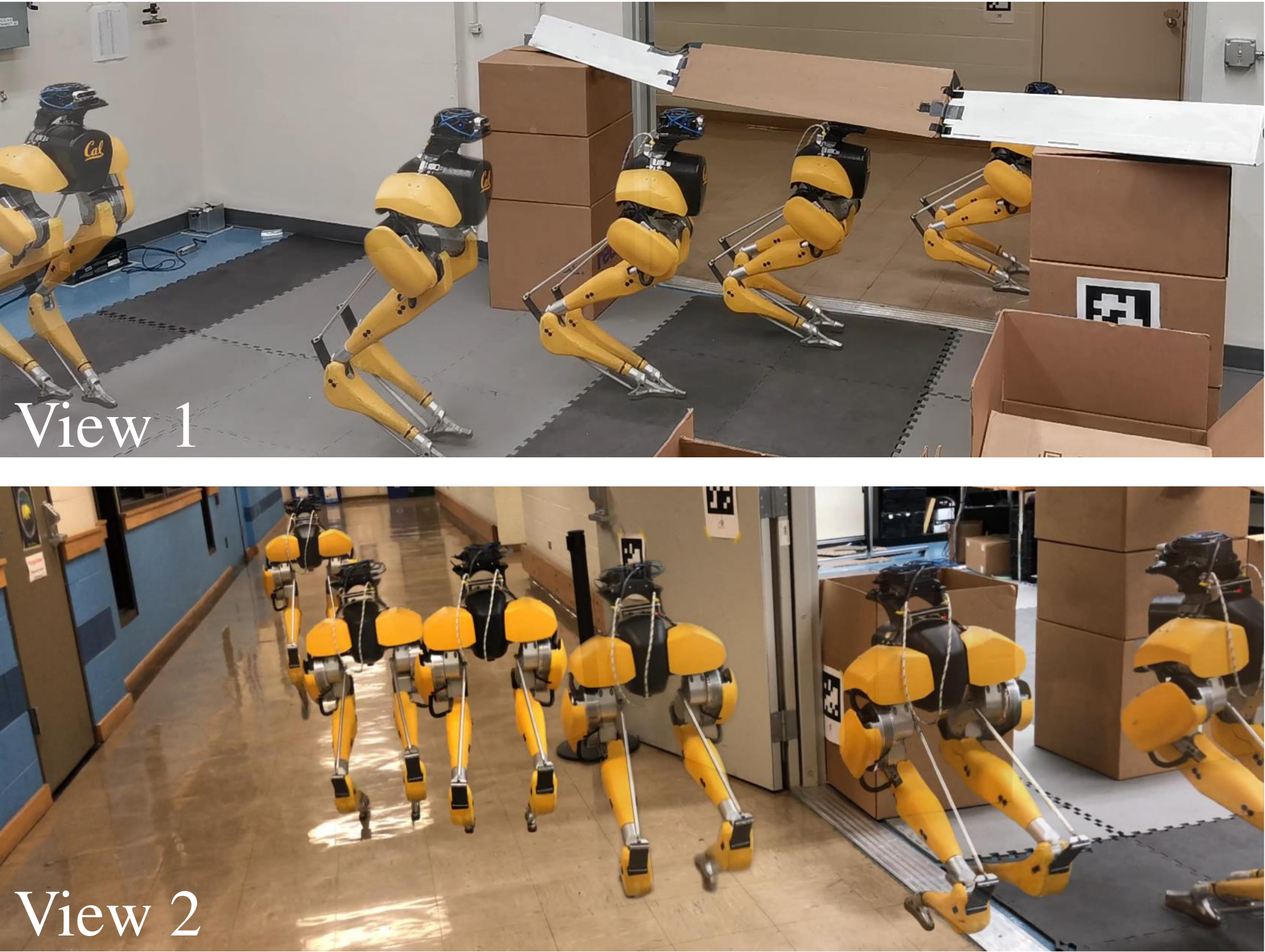}
        \caption{A snapshot of experiment with the obstructed door} \label{subfig:door_expr}
    \end{subfigure} 
    \begin{subfigure}{0.235\textwidth}
        \centering
        \includegraphics[width=\linewidth]{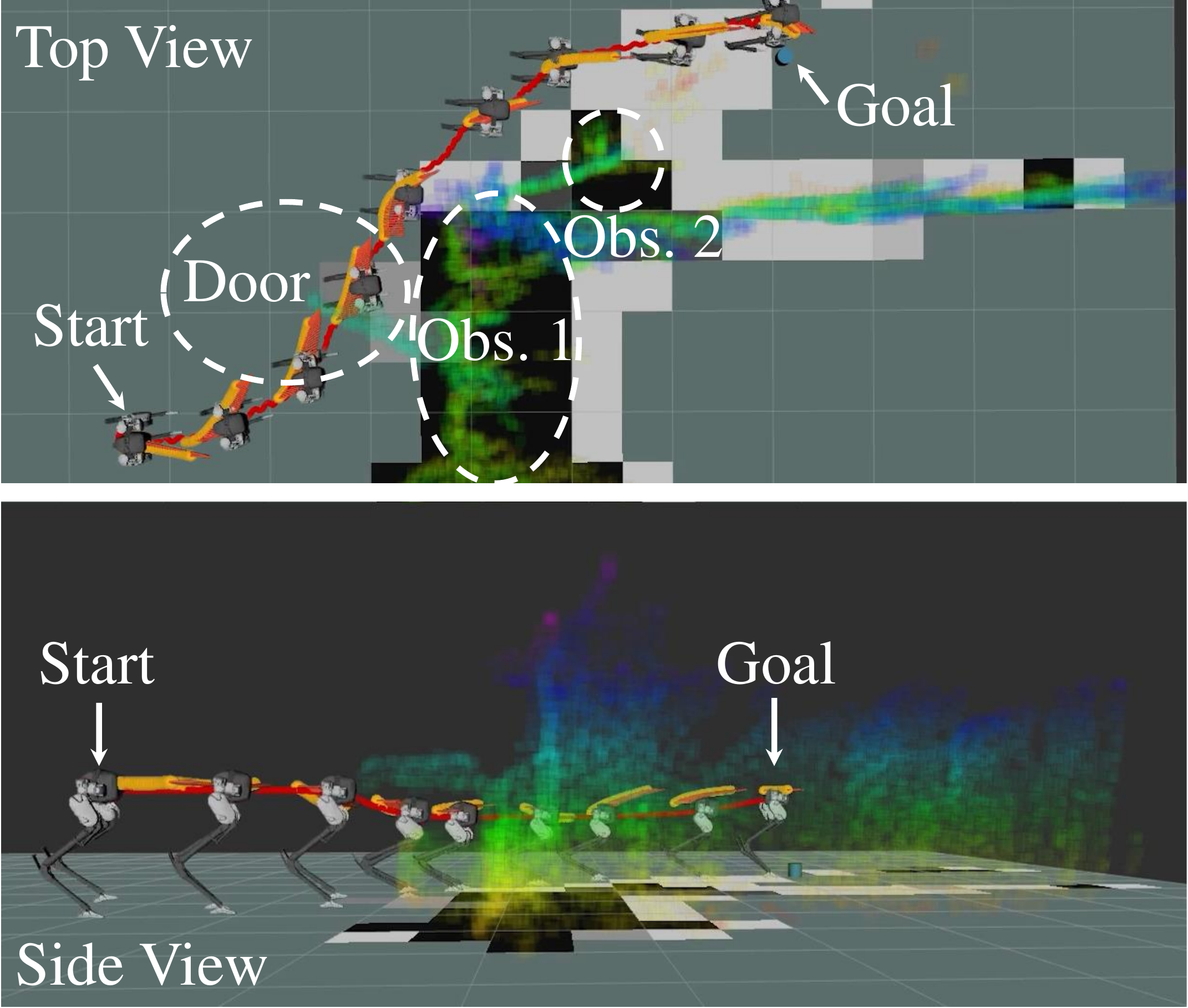}
        \caption{Planning snapshot during the experiment with obstructed door} \label{subfig:door_rviz}
    \end{subfigure} 
    \begin{subfigure}{0.23\textwidth}
        \centering
        \includegraphics[width=\linewidth]{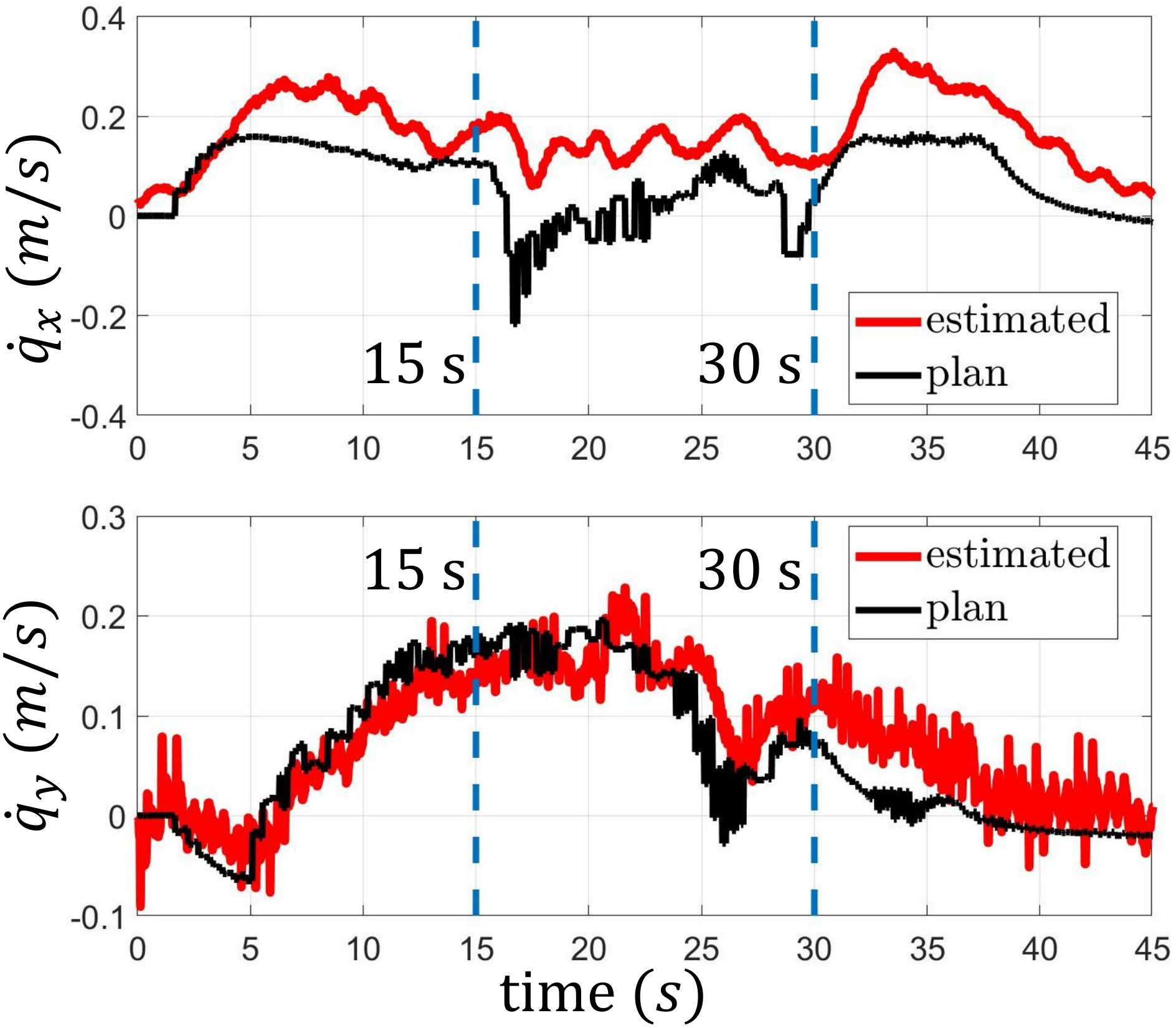}
        \caption{A comparison b/w planned and actual walking speeds} \label{subfig:door_vxvy}
    \end{subfigure}
    \begin{subfigure}{0.245\textwidth}
        \centering
        \includegraphics[width=\linewidth]{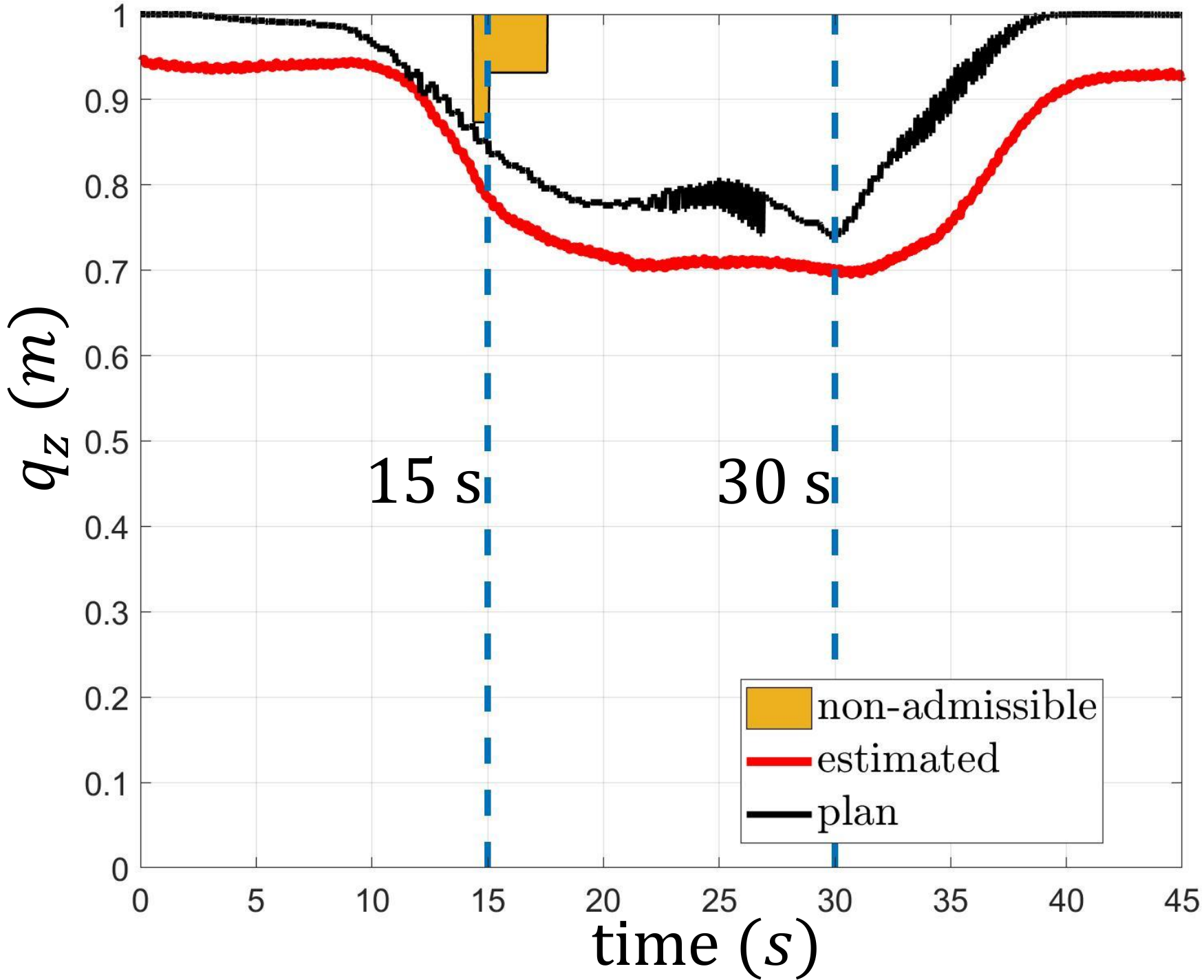}
        \caption{Planned, actual and non-admissible walking height} \label{subfig:door_wh}
    \end{subfigure}
    \caption{Obstructed Door Experiment. An obstacle is hanged in front of a door that is left open, and the robot travels from a lab space (View 1 in Fig.~\ref{subfig:door_expr}) to the corridor outside (View 2 in Fig.~\ref{subfig:door_expr}) through the obstructed door. In the lab space, the robot uses lateral movement and turning to avoid the obstacle~(Obs.~1) and to crouch to walk outside of the lab space. Once the robot perceived the new environment and obstacles in the corridor, such as the stanchion and the opened door circled as an Obstacle~(Obs. 2), it quickly replans to find a new safe path to the goal location.}
    \label{fig:door_result}
    \vspace{-0.2cm}
\end{figure*}

\subsubsection{Height-Constrained Space}
In this test, there is an arch-shaped obstacle with a height of 1~m, causing 0.75~m admissible height, along with two obstacles pinned on the ground. This is shown in Fig.~\ref{subfig:clutter_expr}. 
The robot is initialized right behind the first obstacle in order to have a limited view of the entire map, and the goal is set to be 8~m ahead of the robot's start location, as illustrated in Fig.~\ref{subfig:clutter_rviz}.
After Cassie avoided the first obstacle and saw the overhanging arch obstacle and detected that the admissible height in this area is about 0.75~m, it starts to crouch down to use a walking height of 0.7~m to avoid collision with the ceiling, as illustrated in Fig.~\ref{subfig:clutter_wh}.
Later, Cassie starts to enlarge to its normal walking height to 1~m after it passed through this arch. 
In the meantime, the second obstacle appears in the view of the robot and accordingly, Cassie starts to walk to its right to avoid it and reach its goal location, as shown in Fig.~\ref{subfig:clutter_vxvy}.
Humanoid robots walking underneath similar scenarios are mostly presented in simulation in prior efforts such as~\cite{schulman2014motion,dai2014whole,grey2017footstep}, which use offline optimization with perfect knowledge of the
environment and robot dynamics. 
This is one of the first experimental demonstrations of a person-sized bipedal robot safely navigating in a cluttered environment that has an overhanging obstacle through real-time online planning.


\begin{figure*}[!tp]
\centering
\begin{subfigure}{0.35\textwidth}
    \centering
    \includegraphics[width=\linewidth]{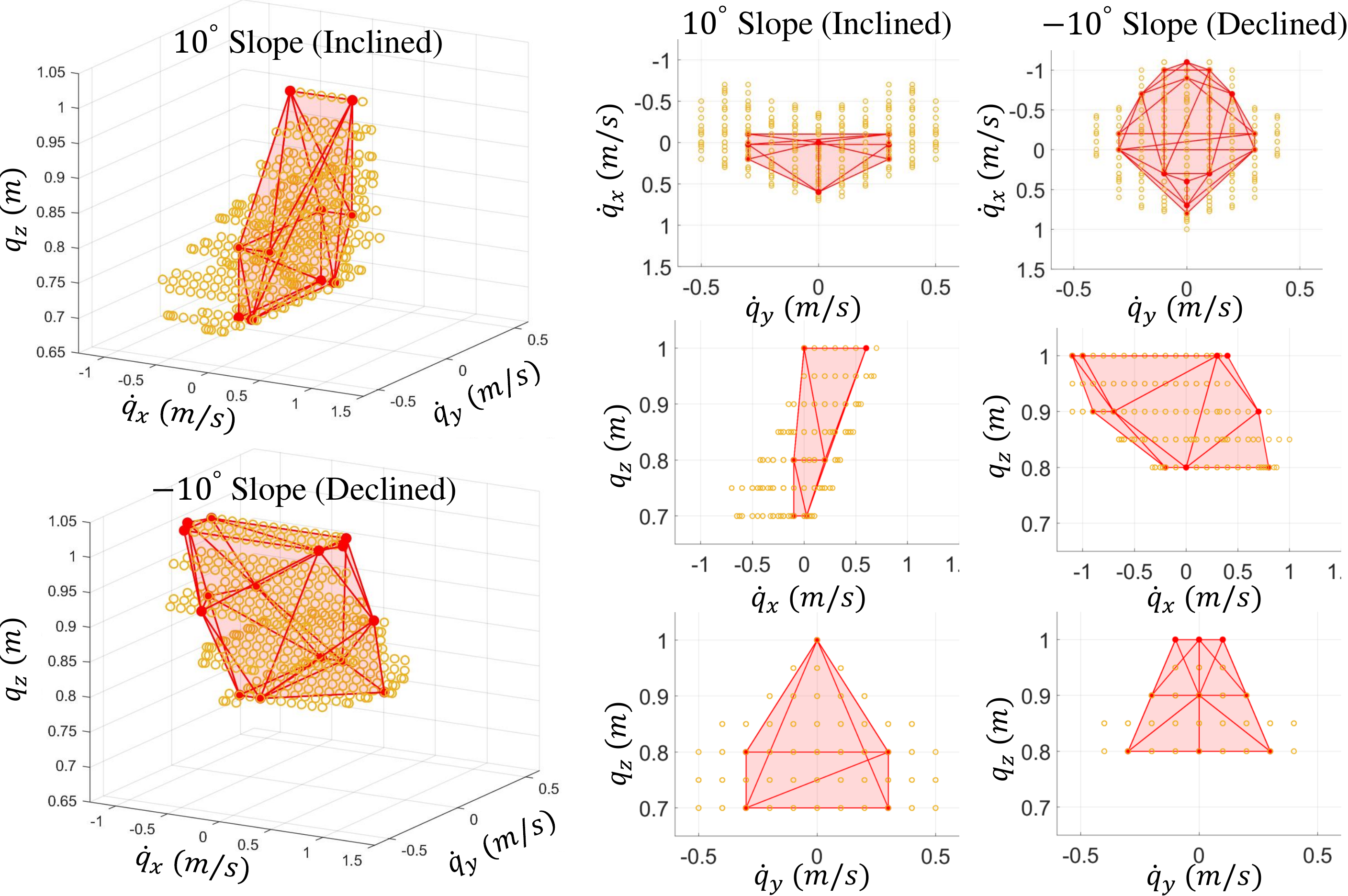}
    \caption{Feasible command sets and their convex feasible command subsets when the robot is walking at different slopes.} \label{subfig:feasbile_cmd_set_slopes}
\end{subfigure} 
\begin{subfigure}{0.39\textwidth}
    \centering
    \includegraphics[width=\linewidth]{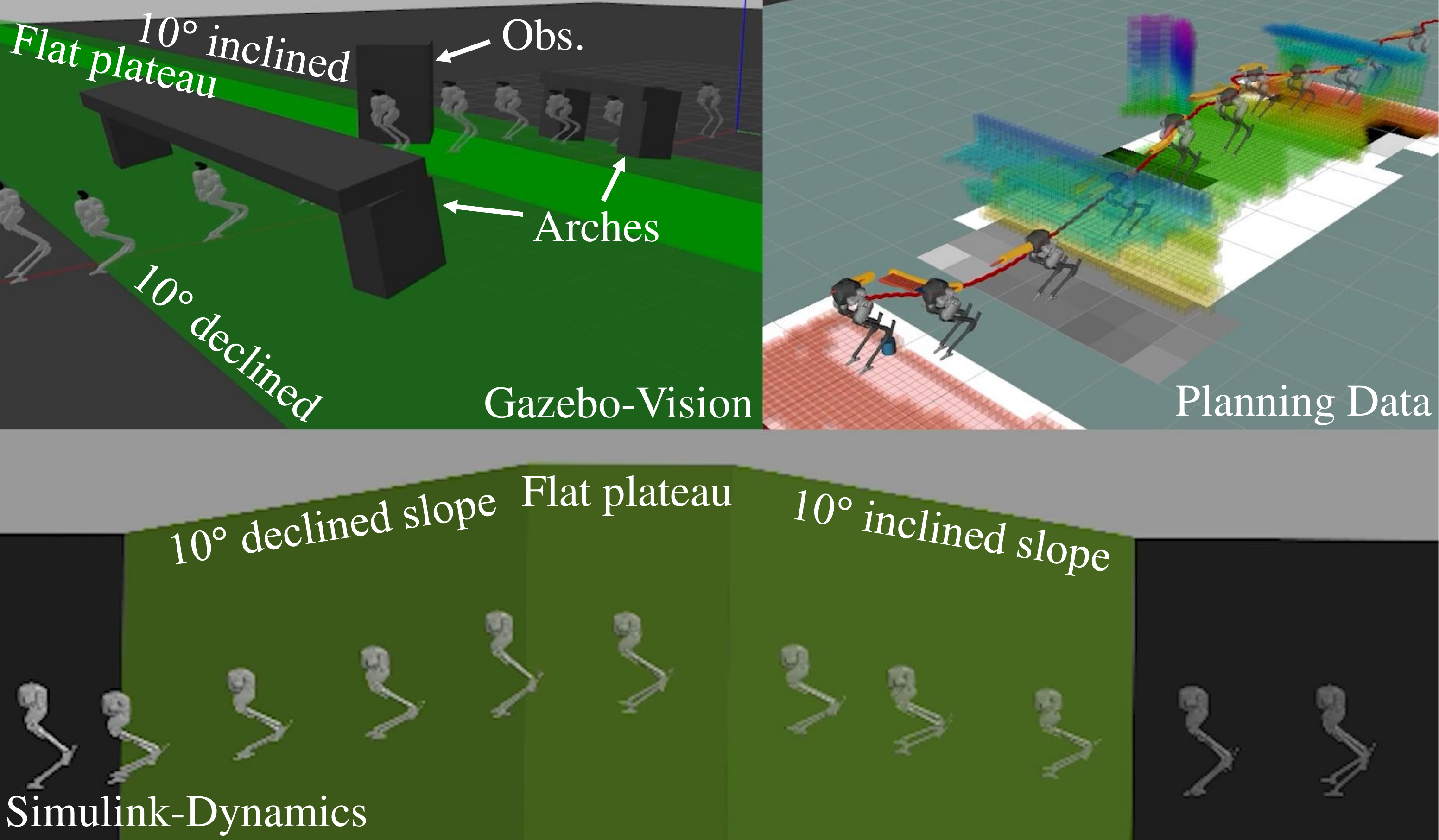}
    \caption{A snapshot of joint simulation of vision and dynamics using the proposed method to navigate the robot on different slopes with height-constrained regions.} \label{subfig:sim_slope_vis}
\end{subfigure} 
\begin{subfigure}{0.24\textwidth}
    \centering
    \includegraphics[width=\linewidth]{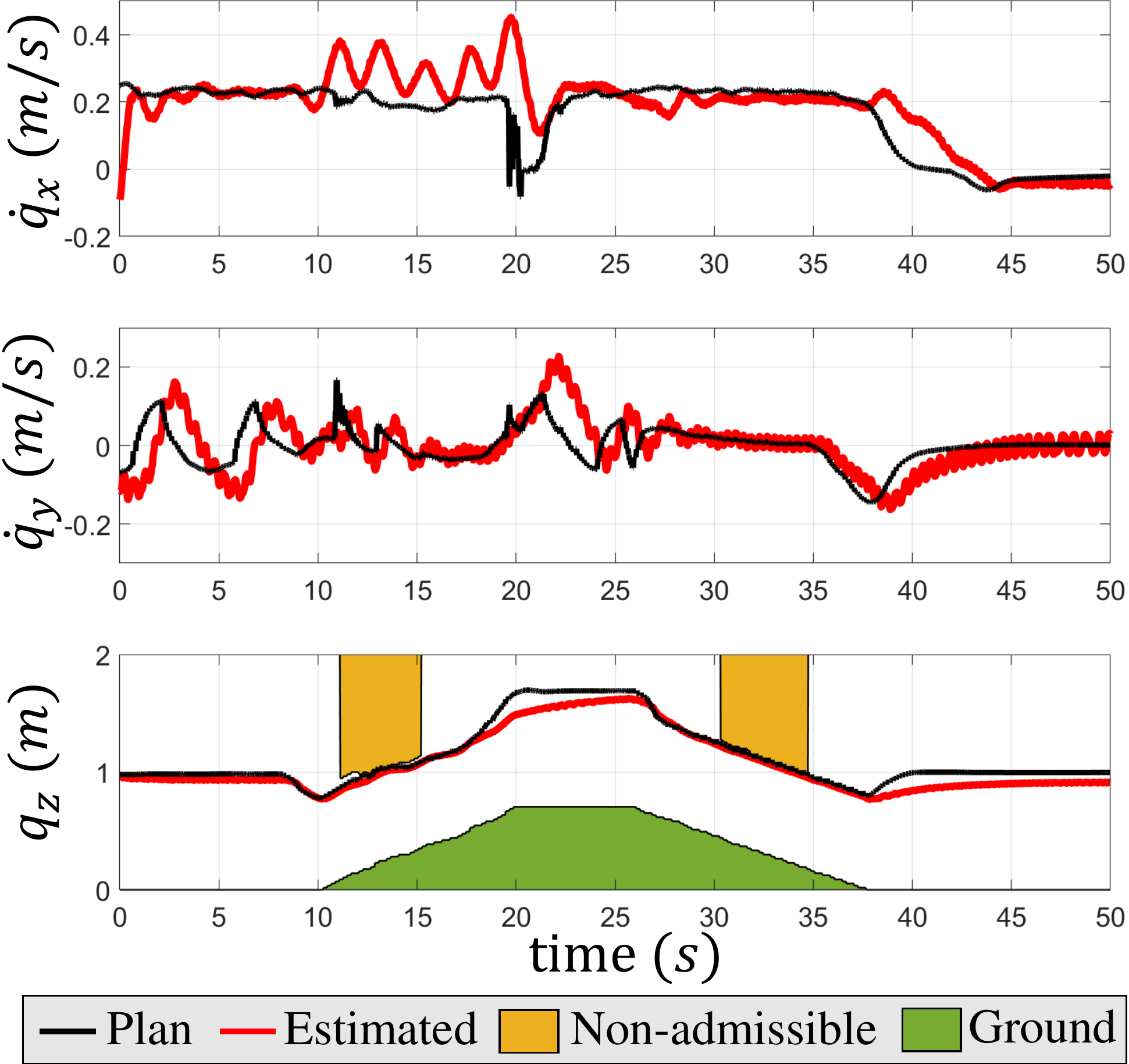}
    \caption{Planned and actual walking speeds and walking height while navigating on sloped terrain.} \label{subfig:sim_slope_plot}
\end{subfigure}
\caption{Extension to navigating the bipedal robot Cassie on sloped terrain with ground obstacles and height-constrained regions. (a) Feasible command sets and convex feasible command subsets when the robot is walking at $10^\circ$ inclined slope and $10^\circ$ declined slope, respectively. The yellow circles represent feasible commands while the red dots are selected vertices to form a convex subset. These sets are obtained in the same way as described in Sec.~\ref{subsec:safeset} by incorporating the slope angle during simulation. (b) The navigation autonomy is validated in a Simulink-Gazebo joint simulation as described in Sec.~\ref{subsec:sim_implementation}. The proposed method enables the robot to safely maneuver through height-constrained regions (arches) on sloped terrains while avoiding obstacles. When the robot is walking on different slopes, the corresponding convex feasible command set is considered during planning. The planning data in this test is recorded in (c). The simulation video is at:~\url{https://youtu.be/h7-uVZUIzvA}.}
\label{fig:slope_sim}
\end{figure*}

\subsubsection{Multiple Arches}
In order to show the autonomy framework is capable to operate in more constrained environments where the robot needs to keep a low walking height, two arches are placed in the two ends of the 4~m$\times$10~m space, respectively, representing the entrance and exit of this constrained area, as demonstrated in Fig.~\ref{subfig:tunnel_expr}.
Moreover, a 0.5~m$\times$0.5~m obstacle is placed inside of this region. 
In the experiments, Cassie changes its walking height to 0.7~m in order to walk through the first arch, starting after 11~s as shown in Fig.~\ref{subfig:tunnel_wh}.
During the robot's travel inside this tight space using a 0.7~m walking height illustrated in Fig.~\ref{subfig:tunnel_rviz}, there are several deceleration happening in order to finely regulate Cassie's position to avoid collision with the walls and obstacles nearby, such as 26~s, 38~s, and 42~s in Fig.~\ref{subfig:tunnel_vxvy}.
Since there are no more overhanging obstacles detected, the planners start trying to recover Cassie's height to a normal height, as shown between 21~s to 27~s in Fig.~\ref{subfig:tunnel_wh}. 
However, at that time, the robot has traveled forward and detected the second arch, therefore, the planners quickly replan to keep the robot at a low walking height.
After the robot exited this area after 50~s, the robot recovers to a normal walking height. 
However, as shown in Fig.~\ref{subfig:tunnel_rviz} and attached video, Cassie passed the goal location while it is enlarging the walking height. 
The planner is then able to correct the robot back to the goal location by sending backwards velocity commands as recorded after 65~s in Fig.~\ref{subfig:tunnel_vxvy}.
This experiment first exhibits a capability of the proposed autonomy for bipedal robots to keep low walking height under multiple height-constrained spaces while reactively avoiding obstacles.

\subsubsection{Obstructed Door}
As demonstrated in Fig.~\ref{subfig:door_expr}, this environment consists of two parts: a lab space that has a size of 4~m$\times$10~m and a corridor that is outside of the lab and has a width of 4~m. 
There is a door left open connecting these two space and an arch-like obstacle is placed in front of the door. 
Such an overhanging obstacle adds height constraint to the doorway. 
The robot is initialized inside the lab and the goal location is set to be outside the lab, as shown in Fig.~\ref{subfig:door_rviz}.
After the robot detected the obstacles and the doorway, the planners start to drive Cassie to its left in order to move towards the target, as shown in Fig.~\ref{subfig:door_vxvy}.
In the meantime, Cassie starts to crouch down to avoid the arch obstacle over the doorway. 
After Cassie moved outside of the lab, obstacles in the corridor, such as the stanchion and the opened door shown in Fig.~\ref{subfig:door_expr}~(View 2), can be detected and robot therefore changes its previous plan and keeps moving to its left to avoid these new-detected obstacles. 
Moreover, when the robot is moving across the doorway, the detected height-constrained grids surrounding the robot are included in the local map, even though the robot is not walking exactly underneath them. 
Therefore, as illustrated in Fig.~\ref{subfig:door_wh}, Cassie keeps a low walking height between 15~s and 30~s in this cluttered environment until 30~s when it travels to an open space and recovers to a normal height.
This shows a successful demonstration of a large-scale humanoid robot walking underneath an obstructed door and traveling between different rooms. 

\section{Extension to Navigation on Height-Constrained Sloped Terrain}~\label{sec:slope}
While all earlier results are conducted on flat ground, in this section, we extend the proposed method to complex terrains with varying slopes. 

In order to realize bipedal navigation on sloped grounds, we require only a few modifications to the navigation framework described in Fig.~\ref{fig:framework}. 
They are: (i) pre-computing the convex feasible command set on different slopes in the same way described in Sec.~\ref{subsec:safeset}, (ii) considering the feasible command sets (by changing the set used in~\eqref{eq:feasible_set_constraint}) based on the current slope angle during planning, and (iii) compensating the terrain heights for the detection of the ground and overhanging obstacles during mapping described in Sec.~\ref{subsec:mapping-details}. We use the same HZD-based variable walking height controller introduced in Sec.~\ref{subsec:controller} as it is robust to the change of ground slope angle in a given range without further tuning or modifications. The controller does not require terrain information. 

Followed by the same method introduced in Sec.~\ref{subsec:safeset}, we obtained the feasible command sets and their convex subset for the Cassie driven by its variable walking height controller on an inclined slope whose angle is $10^\circ$ and on a declined slope with an angle of $10^\circ$. They are shown in Fig.~\ref{subfig:feasbile_cmd_set_slopes}. 
Compared to the feasible command set on the flat ground (zero slope angle) in Fig.~\ref{fig:safe_command_set}, we can clearly observe the set shape and volume changes significantly with the change of the slope angles. 
For example, the feasible command set on the inclined slope shrinks in the backward sagittal velocity dimension $\dot{q}_x$ while the set on the declined slope is smaller in the lower walking height dimension $q_z$. 
This shows that the capacity of the walking controller varies with the changing slope angles and considering such a change during planning is necessary. 
Therefore, the convex feasible command set used in~\eqref{eq:feasible_set_constraint} is updated when solving~\eqref{eq:problem} based on the slope angle at the robot's current footprint. 

To evaluate the modified navigation framework, we build up a simulation environment where the slope angle of the terrain is varying, as shown in Fig.~\ref{subfig:sim_slope_vis}. 
The terrain has a 4-meter-long inclined slope with an angle of $10^\circ$, followed by a 2-meter flat ground, and a 4-meter-long declined slope (slope angle is $-10^\circ$). 
There is an arch on each inclined slope and declined slopes and a ground obstacle on the flat plateau. 
The simulation environment is built in the same high-fidelity Simulink-Gazebo joint simulation of Cassie described in Sec.~\ref{subsec:sim_implementation}. 

As demonstrated in Fig.~\ref{subfig:sim_slope_vis},\ref{subfig:sim_slope_plot}, the proposed navigation framework enables the bipedal robot Cassie to navigate on both inclined and declined slopes while avoiding obstacles and changing the walking height to walk through arches without losing balance or collision. 
In order to reach the target that is set to $13.5$~m ahead of the robot, the robot starts to accelerate after it is initialized on the flat ground. After it walks on the inclined slope, it encounters a height-constrained space, and the robot quickly crouches down and maneuvers through the arch while climbing the slope. After passing the arch, the robot reaches the flat plateau and detected a new ground obstacle. The planner exerts an emerging stop command in front of this obstacle, as recorded around 20 s in Fig.~\ref{subfig:sim_slope_plot}, and the robot takes side steps to move around it. Afterward, the robot walks to the declined slope and successfully reacts to the newly-detected height-constrained space. The robot reaches the goal location and gets back to the normal walking height after passing this declined slope with the overhanging obstacle.  

A limitation of the current method is that the terrain slopes are assumed to be perfectly known and extracting both overhanging obstacles and terrain height on a long slope could be challenging onboard. 
This drawback from our current robot vision algorithm stops us from deploying the system in the real world. Additional efforts on terrain perception by fusing the IMU reading, the robot's leg kinematics, and vision, or a better sensor such as long-range 3D LiDAR, can facilitate the real-world deployment of this navigation autonomy.

For other types of rough terrain such as stairs, we hypothesize that the proposed method could also be flexible for the following reasons. 
On the discrete terrain, the feasible command set can now be formulated to capture the coupled walking dynamics in step length, step width, and step height dimensions, and the rest of the framework can be used. 
But this may require the development of a precise foot placement controller for a 3D bipedal robot extended from the work like~\cite{nguyen2018dynamic}, which is challenging.

\section{Conclusion and Future Work} \label{sec:conclusion}
We have presented one of the first autonomous navigation frameworks for person-size bipedal robots to travel and explore unknown height-constrained and cluttered environments while maintaining gait stability and safety.
The autonomy framework is deployed and validated on an underactuated bipedal robot Cassie to travel in various kinds of congested environments.
This autonomy leverages a lightweight perception system that only includes an RGB-Depth camera and a tracking camera to perceive the world, along with hierarchical real-time planners. 

A global planner using A* search is able to find and reroute the path from the robot's current position to the goal location while considering the cost of different types of areas while exploring an unknown environment.
In order to find a dynamically feasible trajectory for the bipedal robot to track the local goal from the global planner, two optimization-based trajectory planners that consider the robot's physical limitation, gait stability, and obstacle avoidance are developed for real-time replanning. 
In this trajectory planning schematic, a local planner schedules a trajectory to reach the given local goal and a reactive planner considers the real-time commands for the variable walking height controller to follow the planned trajectory.

Moreover, a vertical-actuated Spring-Loaded Inverted Pendulum~(vSLIP) is developed to capture the coupled walking dynamics of the bipedal robot in the varying walking heights scenarios during trajectory optimization. 
We introduce a new and practical method to develop and utilize a reduced model (\textit{i.e.}, vSLIP) by first investigating the walking patterns and restrictions on the robot controlled by an HZD-based walking controller, and later utilizing such information in the reduced model for online optimization.

Empowered by the proposed autonomous navigation framework, we demonstrate several successful experiments of Cassie autonomously exploring a diverse repertoire of height-constrained narrow environments, such as a cluttered space with an overhanging obstacle, a region that has multiple arches, and an obstructed door.
During the experiments, Cassie exemplifies its capacity to reactively avoid obstacles on the ground and to crouch down and travel underneath arches. Such skills were only realized in simulation with the motions pre-computed offline in most previous work for humanoid robots.
Besides allowing the robot to crouch under a minimum ceiling height of 1.0~m overhanging obstacle, the proposed walking autonomy also exhibits the ability to safely walk through a narrow space that is as wide as the robot's footprint, such as the experiment in the 2D maze. 
Using the proposed vSLIP model and cascading planning strategy, Cassie never loses its gait stability, \textit{i.e.}, never falls over, throughout the tests of this work.

We next outline a few directions for future work. In our work, we use a single RGB-Depth camera for obstacle detection, enabling the robot to have a small-sized sensor stack and thus enabling walking underneath low-hanging obstacles. However, the field of view of a single depth camera is limited, and therefore, the robot cannot take a wider range of environments into consideration, such as the obstacles behind the robot. 
This can be solved by upgrading to a 3D LiDAR and this is one of our future work. 
Moreover, the variable walking height controller used in this work shows some delays and tracking errors not only in the walking velocity but also in the walking height. This forces us to include safety margins during planning and further limits the agility of this autonomy. 
Replacing the walking controller in this pipeline with a more powerful one with better control performance is another future work.
Furthermore, Cassie is able to maintain a lower walking height of 0.6~m by using the current walking controller, which can further extend its capacity in tighter spaces.
However, Cassie cannot reach that limit in the presented experiments because the state estimation by the tracking camera is no longer stable due to large vibrations from faster stepping frequency and impacts at such a low walking height. We addressed this by setting a minimum walking height of $0.7$~m during experiments.
Investigating a reliable state estimator for bipedal robots that can handle consistent and large impacts at varying walking heights is a very interesting research direction.

\section*{Acknowledgments}
This work is supported in part by National Science Foundation Grant CMMI-1944722. The authors would also like to thank Bike Zhang, Ayush Agrawal, and Lizhi Yang for their gracious help in experiments.

\balance
\bibliography{references}{}
\bibliographystyle{IEEEtran}
\end{document}